\documentclass[lettersize,journal]{IEEEtran}
\pdfoutput=1
\usepackage{amsmath,amsfonts}
\usepackage{algorithmic}
\usepackage{algorithm}
\usepackage[ruled,algo2e]{algorithm2e}
\usepackage{array}
\usepackage{textcomp}
\usepackage{stfloats}
\usepackage{url}
\usepackage{verbatim}
\usepackage{subfigure}
\usepackage{graphicx}
\usepackage{cite}
\def\BibTeX{{\rm B\kern-.05em{\sc i\kern-.025em b}\kern-.08em
    T\kern-.1667em\lower.7ex\hbox{E}\kern-.125emX}}
\usepackage{balance}
\begin{document}
\title{Learning in a Single Domain for Non-Stationary Multi-Texture Synthesis}
\author{Xudong Xie, Zhen Zhu, Zijie Wu, Zhiliang Xu, Yingying Zhu
\thanks{Corresponding author: Yingying Zhu. 

X. Xie, Z. Wu, and Y. Zhu are with the School of Electronic Information and Communications, Huazhong University of Science
and Technology, Wuhan, 430074, China. E-mail: \{xdxie, zijiewu, yyzhu\}@hust.edu.cn.

Z. Xu is with Baidu Inc., Shenzhen, 518062, China. E-mail: xuzhiliang@baidu.com.

Z. Zhu is with the Department of Computer Science, University of Illinois at Urbana-Champaign, Champaign, IL, 61820. E-mail: zhenzhu4@illinois.edu.

}}

\maketitle

\begin{abstract}
This paper aims for a new generation task: non-stationary multi-texture synthesis, which unifies synthesizing multiple non-stationary textures in a single model. Most non-stationary textures have large scale variance and can hardly be synthesized through one model. To combat this, we propose a multi-scale generator to capture structural patterns of various scales and effectively synthesize textures with a minor cost. However, it is still hard to handle textures of different categories with different texture patterns. Therefore, we present a category-specific training strategy to focus on learning texture pattern of a specific domain. Interestingly, once trained, our model is able to produce multi-pattern generations with dynamic variations without the need to finetune the model for different styles. Moreover, an objective evaluation metric is designed for evaluating the quality of texture expansion and global structure consistency. To our knowledge, ours is the first scheme for this challenging task, including model, training, and evaluation. Experimental results demonstrate the proposed method achieves superior performance and time efficiency. The code will be available after the publication.
\end{abstract}


\section{Introduction}

\IEEEPARstart{T}{exture} synthesis, as one of the classical low-level computer vision tasks, has broad application prospects in image manipulation~\cite{shaham2019singan}, image restoration~\cite{bertalmio2003simultaneous}, 3D reconstruction~\cite{zeng2022joint}, and so on. 
% As a sub-task of texture synthesis, texture expansion can be defined as generating an expanded image in which more texture elements are created according to the original arrangement and structure.
Current texture synthesis methods can be roughly divided into single-texture synthesis (STS) and multi-texture synthesis (MTS). Fig.~\ref{fig:intro} (a) and (b) illustrate the difference.
STS trains a model for each texture image separately. 
% The former~\cite{gatys2015texture,jetchev2016texture,zhou2018non,Shocher_2019_ICCV,shaham2019singan} trains a model for each texture image separately, while the latter trains one model for multiple texture images.
Though some methods perform well on either a single natural image~\cite{shaham2019singan,hinz2020improved} or non-stationary texture image~\cite{zhou2018non}, they are all time-consuming since for each instance, they need to train a separate model for up to several hours. MTS methods are proposed to solve this and are trained on texture datasets by designing new training strategies~\cite{li2017diversified} and network structures~\cite{shi2020fast}, or transforming the distribution of feature maps~\cite{li2017universal,yu2019texture}. Nevertheless, most MTS methods cannot deal with irregular and non-stationary textures, as shown in Fig.~\ref{fig:intro} (d). Moreover, global structures of the generated textures are severely distorted.

{
\begin{figure}[tb]
\centering
\subfigure[]{\includegraphics[width = 0.22\textwidth]{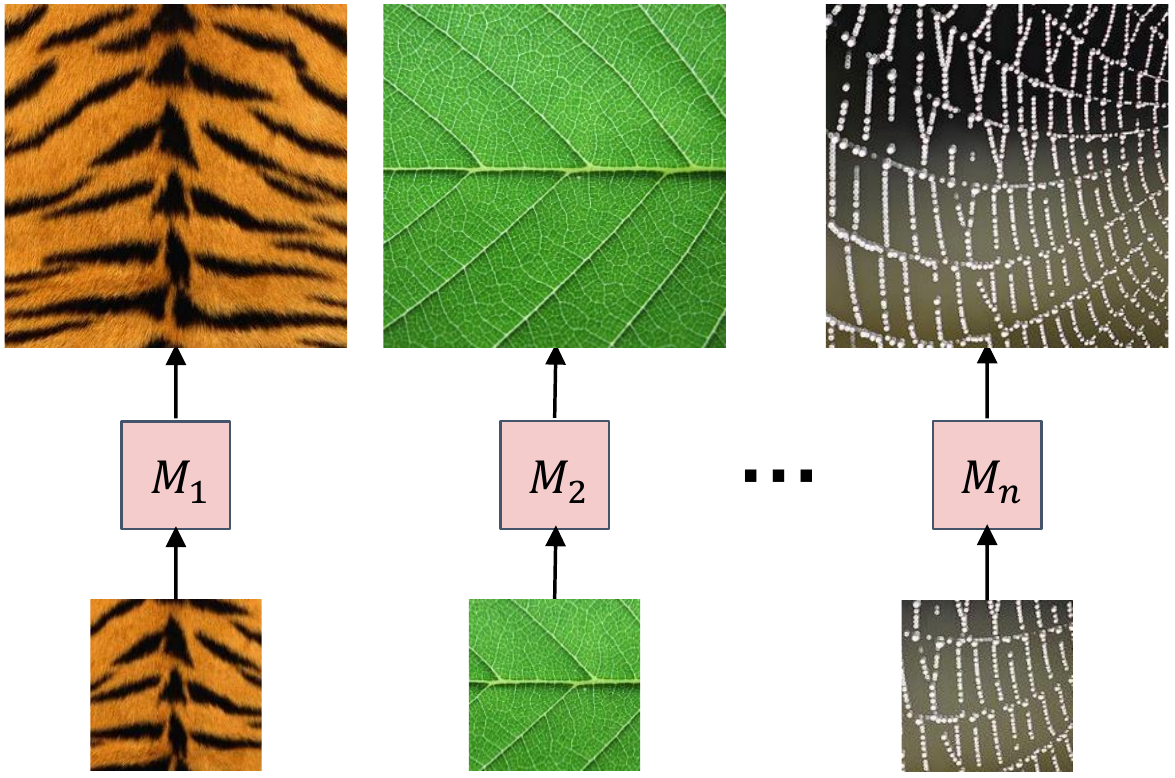}}
    \label{fig:short-a}
\subfigure[]{\includegraphics[width = 0.22\textwidth]{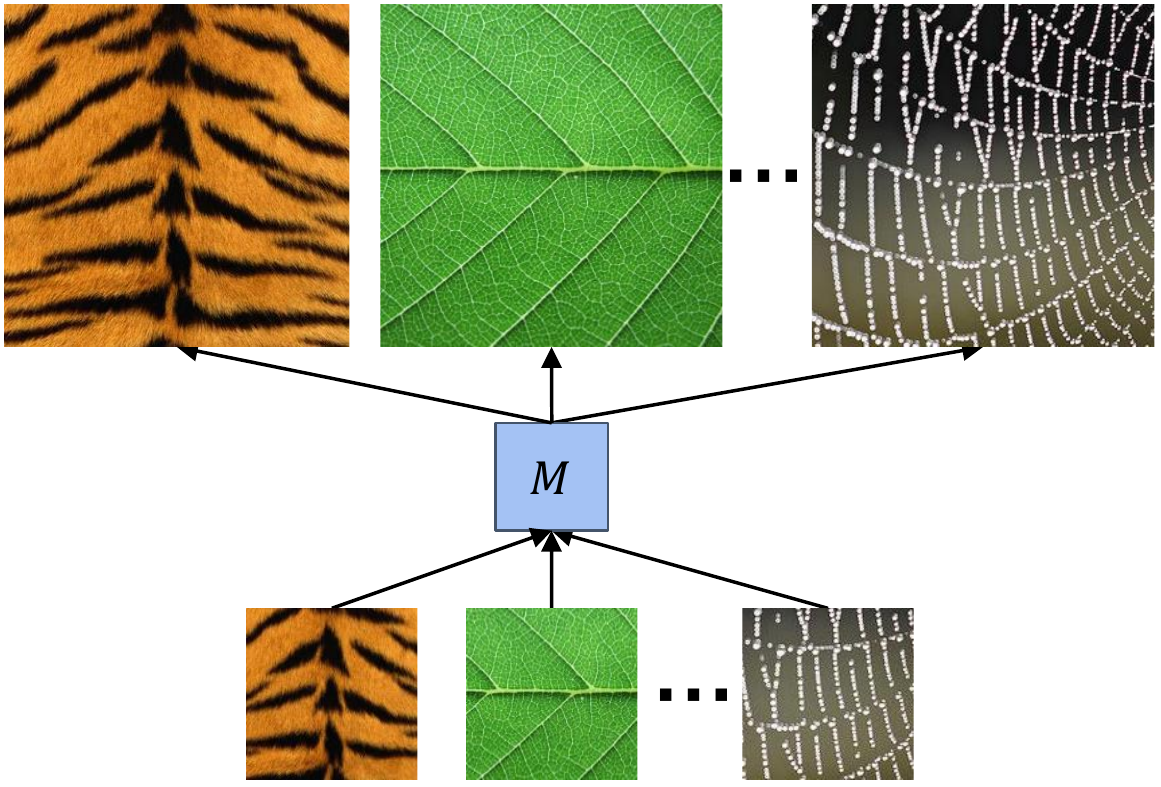}}
    \label{fig:short-b}
  
\subfigure[]{\includegraphics[width = 0.22\textwidth]{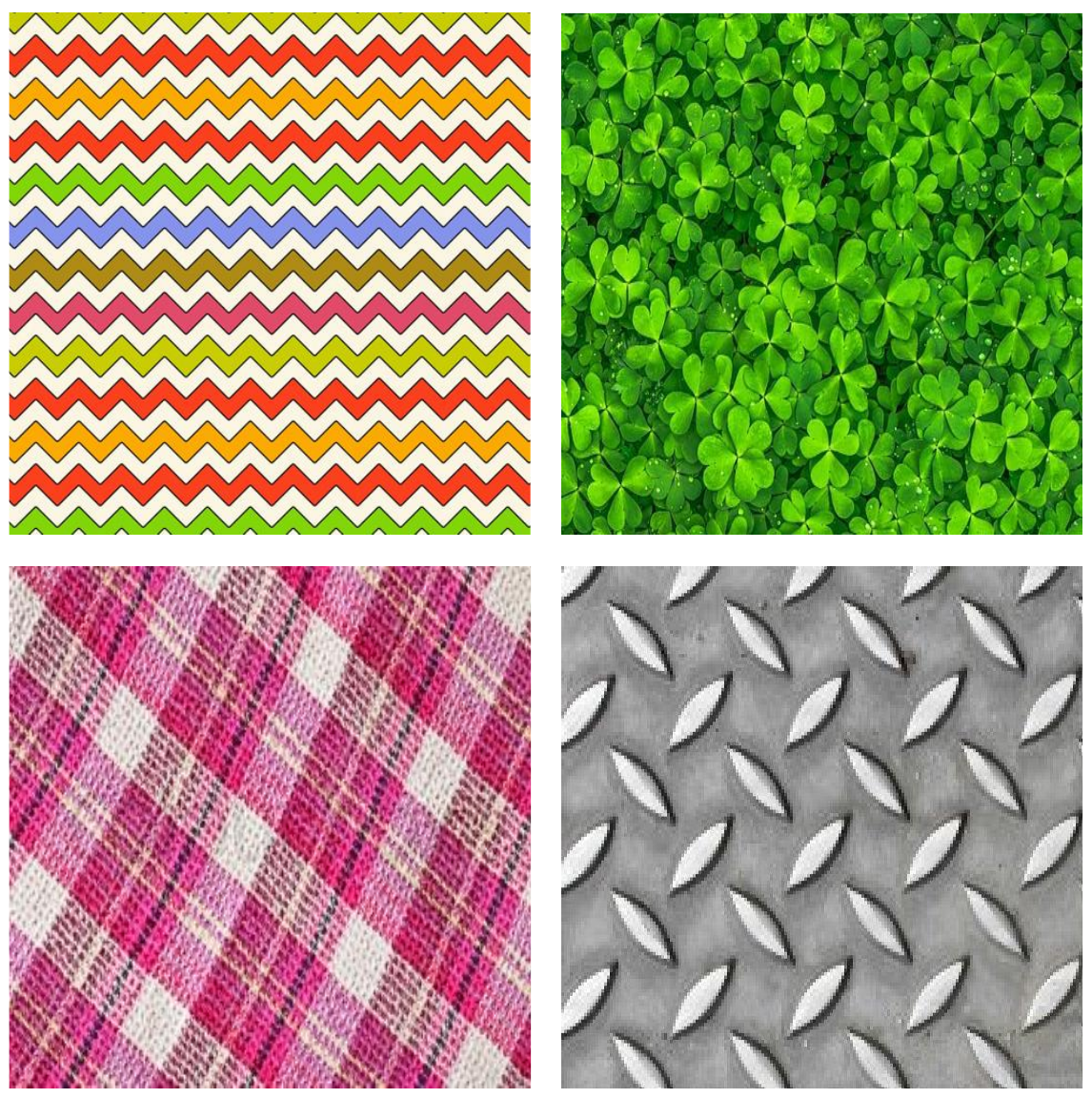}}
    \label{fig:short-c}
\subfigure[]{\includegraphics[width = 0.22\textwidth]{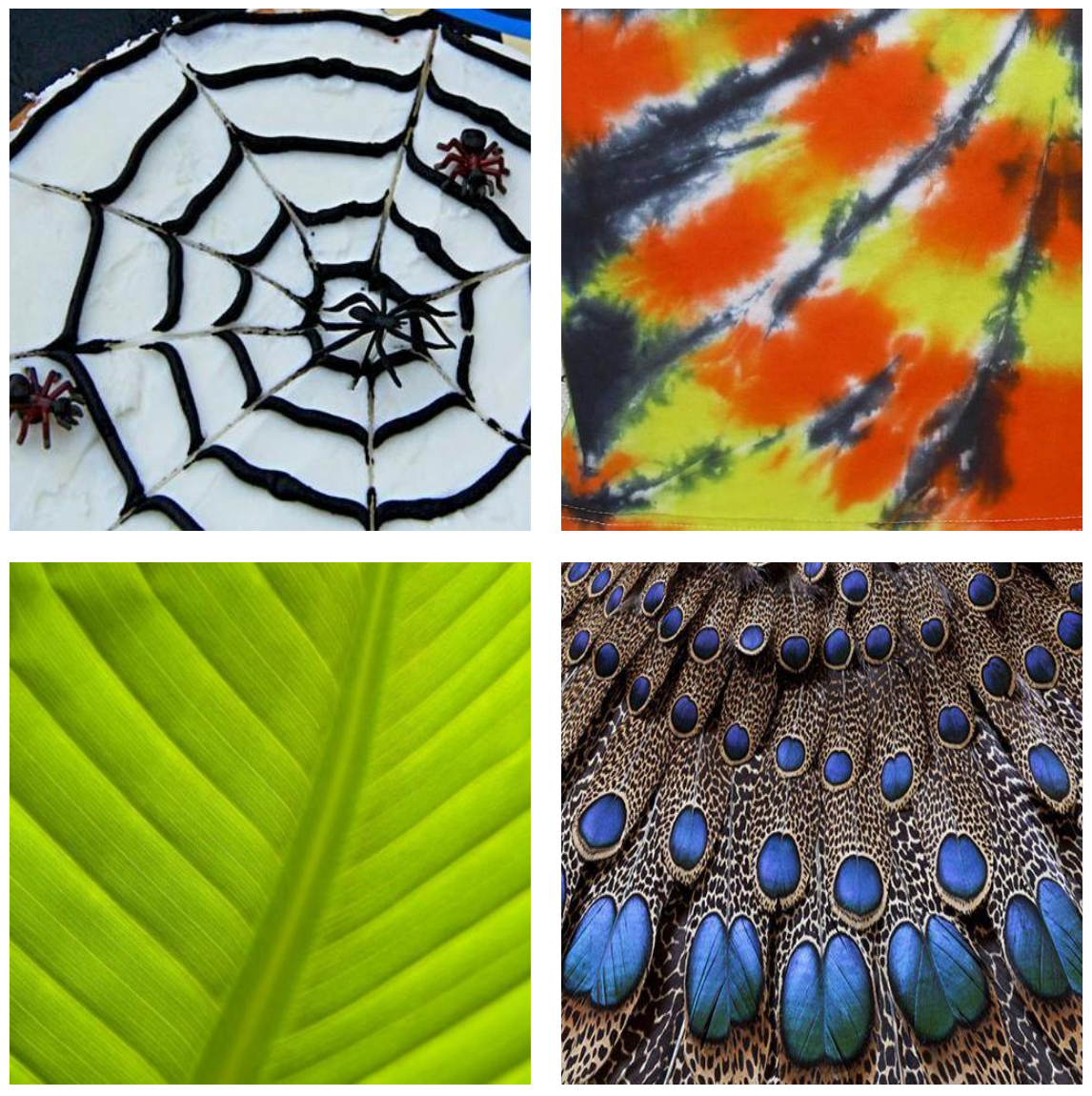}}
    \label{fig:short-d}
    
\caption{(a) and (b) illustrate the difference between the methods of STS and MTS. (c) and (d) demonstrate several stationary and non-stationary textures separately, where the latter are usually with anisotropic large-scale global structure, complex background, and spatial appearance variance, for instance, cobwebbed, veined, etc.}
\label{fig:intro}
\end{figure}
}

% Figure~\ref{fig:intro} demonstrates several stationary and non-stationary textures, where the latter are usually with anisotropic large-scale global structure, complex background, and obvious spatial appearance variance, for instance, cobwebbed, veined, stained, cracked and so on.

\begin{figure*}[t!]
	\centering
	\includegraphics[width=0.95\textwidth]{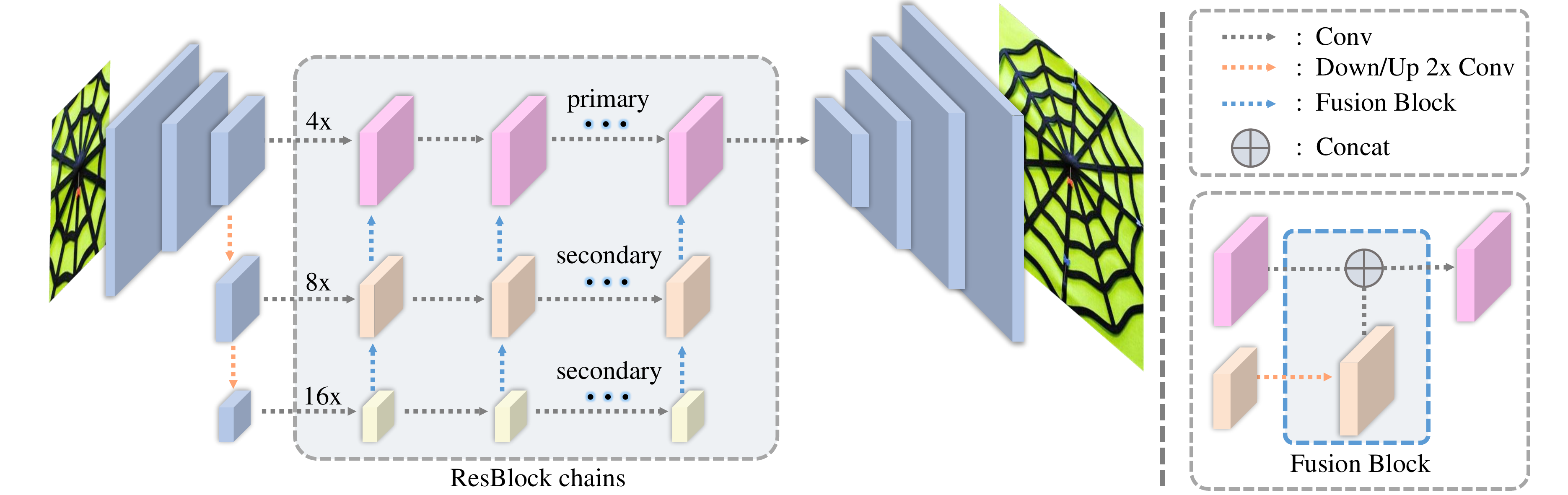}
	\caption{Generator architecture of MTGAN. The three branches correspond to 4$\times$, 8$\times$ and 16$\times$ down-sampling respectively. The input and output sizes are $128\times 128$ and $256\times 256$ in the training phase. By default, we use the three-branch model to generate the final results. }
    \label{fig:framework}
\end{figure*}

Based on the above observation, we focus on a challenging and meaningful task: non-stationary MTS in a single domain. That is, training a model to expand multiple different non-stationary textures of a single category while also preserving the global structure of the original textures. One noticeable benefit of this task is the well-trained models are able to produce multi-pattern variations of non-stationary textures that can not be observed in STS methods. Yet, one challenging aspect of this task is various scales of structures may appear in different samples or even in a single sample for non-stationary textures. In view of this, it is necessary to design a model with solid scale sensitivity to capture texture information of different scales. 
% We argue multi-scale deep network architectures~\cite{snelgrove2017high,wang2020deep} are helpful because they enhance the network's representation power for different structure scales.
Therefore, we design a multi-scale generator by building multiple parallel network branches with different down-sampling rates, as shown in Fig.~\ref{fig:framework}. 
% The primary branch fuses multi-scale information from several secondary branches block by block and branch by branch. Through these strategies, 
The generator can preserve the global structure and generate meaningful and real-world texture images by training on multiple samples.

% On the other hand, we find that the global structure similarity is high among textures of the same category. 

Besides, we observe that regularly, textures of the same category show similar structural patterns, \textit{e.g.}, cobweb textures usually show the same radial gradient expansion pattern.
% Therefore, each sample can obtain global guidance information from other samples, which can reduce the difficulty of model learning and make it converge to a class center. 
This hints to us that training on different samples in the same category benefits from the complementary information of each other. Otherwise, if training the model on samples from different categories, it is likely to have much longer training time and greater learning difficulty.
% But this phenomenon does not appear in samples from different categories. 
Therefore, we propose a simple and effective training strategy, that is, category-specific training rather than random training on different categories. We prove that the model can learn the structures and expansion patterns of a specific texture category reasonably well under this scheme. More importantly, once a model is trained on a class of textures, it can be extended to arbitrary unseen images of the same category by a fast fine-tune phase since the pre-trained model is already knowledgeable about basic strategies to expand the inputs. 
We can also use a continual learning-based training method so that the final model can handle several categories at the same time.

In search of metrics for evaluating non-stationary multi-texture synthesis methods, we found regularly used metrics are not suitable to fully demonstrate the actual performance of different methods.
% During our search for reasonable metrics to evaluate the task of non-stationary multi-texture expansion, 
% Currently, there is no uniform objective evaluation metric specifically for texture synthesis tasks and the existing metrics do not perform well. 
To mitigate this problem, we contribute a metric named Multi-Scale Texture Similarity (MSTS). It is calculated upon the distribution distance between multi-scale deep features extracted from a pre-trained texture recognition model. In this way, it can not only evaluate whether the generated images preserve the global structure, but also evaluate the quality of texture synthesis. This is a multi-scale metric that considers the overall global structure and meanwhile focuses on local texture details. Its effectiveness is verified through our user study.

In summary, the contributions can be listed as follows:
% \vspace{-1ex}
\begin{enumerate}
\item We systematically introduce a new generative task: non-stationary multi-texture synthesis in a single domain, including manifesting several applications and potential difficulties of this task.
% multi-scale generator for a very challenging task: non-stationary MTS, to expand textures effectively without destroying the global structures.
% \vspace{-1ex}
\item We propose an effective scheme for this task, which includes a multi-scale generator, a category-specific training strategy, and a texture similarity metric. Quantitative and qualitative results show the superiority of our method in image quality and time efficiency.
\end{enumerate}

% \begin{enumerate}
% \item[1)] We propose a multi-scale generator for a very challenging task: non-stationary multi-texture synthesis, to expand textures effectively without destroying the global structures.
% \item[2)] We present a category-specific training strategy, enabling the model to quickly expand unseen images of the same category, as well as other categories using an incremental learning strategy.
% \item[3)] We design a reasonable evaluation metric specifically for the non-stationary multi-texture synthesis task. Its effectiveness is verified through the user study.
% \end{enumerate}

\section{Related work}

Conventional approaches of example-based texture synthesis have achieved fine results on stationary textures. Image melding~\cite{darabi2012image} presents a patch-based optimization method for synthesizing a gradual transition region between two images. Self-tuning~\cite{kaspar2015self} was proposed as a non-parametric texture optimization algorithm via weighted guidance channels. 
%Utilizing guidance channels and a direction field, Zhou~\emph{et al.}~\cite{zhou2017analysis} implemented the user-control on non-stationary and inhomogeneous textures. 
Recently, deep learning methods have emerged because of their excellent ability to extract meaningful high-level and low-level features. We mainly introduce these from two aspects:

\noindent\textbf{Single-Texture Synthesis.}
Gatys~\emph{et al.}~\cite{gatys2015texture} propose a seminal optimization-based method for this task using CNN. They optimize the input by matching the Gram matrices extracted by VGG network~\cite{simonyan2014very} between the generated image and the reference image. Liu~\emph{et al.}~\cite{liu2016texture} improve the CNN-based training method by incorporating the Fourier spectrum constraints, making the output images preserve large quasi-periodic structures. Rodriguez~\emph{et al.}~\cite{rodriguez2019automatic} generate regular repeatable textures by finding the minimal repeating pattern in single images.
Spatial generative adversarial network (SGAN)~\cite{jetchev2016texture} is the first GAN~\cite{goodfellow2014generative}-based texture synthesis method, treating input noise as a whole spatial tensor.
 %, which can generate plausible textures with arbitrary sizes. 
To handle non-stationary textures, Zhou~\emph{et al.}~\cite{zhou2018non} present another GAN-based model for texture expansion via self-supervised training. Their method is effective for challenging textures with large-scale global structures. InGAN~\cite{Shocher_2019_ICCV} improves this method by inserting a geometric transformation layer into the generator, enabling the model to produce outputs of different sizes and shapes. TileGAN~\cite{fruhstuck2019tilegan} combines smaller resolution outputs of GANs to form a large-scale non-homogeneous texture. Recently, SinGAN~\cite{shaham2019singan} proposes multi-stage generators trained on a single natural image, learning the internal distribution of patches within an image. Some follow-up works, such as ConSinGAN~\cite{hinz2020improved}, SinIR~\cite{yoo2021sinir}, PetsGAN~\cite{zhang2022petsgan} are devoted to speed up the learning efficiency of SinGAN~\cite{shaham2019singan}. They can do well for images with independent objects but fail for global non-stationary structures such as those in Fig.~\ref{fig:single_comparison}. Besides, Darzi~\emph{et al.}~\cite{darzi2022co} capture local texture properties by computing a co-occurrence tensor. Zhou~\emph{et al.}~\cite{zhou2023neural} compute the Guided Correspondence Distance to improve the CNNMRF model for high-quality texture optimization.
However, it should be noted that all single-texture models are time-consuming and costly. Therefore, some works studied how to process multiple textures within one model. Our work is exactly to cope with multi-texture synthesis.

% \begin{figure}[t!]
% 	% \vspace{-0.5cm}
% 	\centering
% 	\includegraphics[width=0.48\textwidth]{figs/framework.pdf}
% 	\caption{Generator architecture of MTGAN. The three branches correspond to 4$\times$, 8$\times$ and 16$\times$ downsampling respectively. Each resblock chain consists of six identical residual blocks. The input and output sizes are $128\times 128$ and $256\times 256$ by default in the training phase. }
%     \label{fig:framework}
% \end{figure}

% However, it should be noted that all single-texture models are time-consuming and costly. Therefore, some works studied how to process multiple textures within one model. Our work is exactly to cope with multi-texture synthesis.

\noindent\textbf{Multi-Texture Synthesis.}
Bergmann~\emph{et al.} extend their SGAN~\cite{jetchev2016texture} as PSGAN~\cite{bergmann2017learning}, allowing the model to process multiple periodic textures concurrently. Meanwhile, Li~\emph{et al.}~\cite{li2017diversified} introduce an incremental learning strategy so as to deal with multiple textures within one model. Other efforts performed MTS by adaptively editing intermediate feature maps of the encoder. WCT~\cite{li2017universal} is a typical method of universal texture synthesis through whitening and coloring operations on features, but the generated images destroy the global structures of input textures according to Fig.~\ref{fig:multi-comparison}. Transposer~\cite{liu2020transposer} treats the encoded feature map as transposed convolution filters and the self-similarity map as the corresponding input. Mardani~\emph{et al.}~\cite{mardani2020neural} view texture synthesis as local Fast Fourier Transform upsampling and performed FFT in feature space for universal texture synthesis. 
Texture mixer~\cite{yu2019texture} is trained on datasets, but its three latent-space operations destroy the global structure. In addition, Shi~\emph{et al.}~\cite{shi2020fast} propose a pseudo optimizer to unfold the iterative optimization procedure of DeepTexture~\cite{gatys2015texture} into a feed-forward neural network, realizing fast texture synthesis. 
% Gao~\emph{et al.}~\cite{9427254} produced texture wallpaper based on multi-label semantic attributes.
However, these approaches cannot handle \emph{non-stationary} textures according to our observation in Fig.~\ref{fig:multi-comparison} and the statement in their paper.
In view of the problems, we present a multi-branch generator for MTS to cope with non-stationary textures effectively.

\noindent\textbf{Image Quality Assessment.}
It remains a problem that how to evaluate the effect of texture expansion and whether the global structure is preserved in the process of texture synthesis. The mainstream full-reference image quality assessment (IQA) models~\cite{ding2021comparison} can be roughly divided into pixel-level methods~\cite{wang2009mean}, structural similarity methods~\cite{wang2003multiscale,wang2004image} and deep-feature based methods~\cite{heusel2017gans,ding2020image,zhang2018unreasonable}. MSE~\cite{wang2009mean} calculates the distance among pixel pairs and SSIM~\cite{wang2003multiscale} separates image structure from brightness and contrast, but sometimes they are inconsistent with the subjective evaluation. Deep-feature based methods, ~\emph{eg.} FID~\cite{heusel2017gans} and DISTS~\cite{ding2020image} extract image features via a pre-trained classification network and then define some operations on them to capture the semantic and structural information. 
In addition, Wu~\emph{et al.}~\cite{wu2018automatic} propose the global and local textureness assessments to automatically guide better texture exemplar extraction for texture synthesis input.
% Recently, some efforts proposed objective metrics for several special visual scenes such as night-time images~\cite{yang2023ehnq} and underwater images~\cite{9913502}.
However, as shown in Fig.~\ref{fig:user_study}, none of these methods are competent for non-stationary multi-texture expansion tasks.

\section{Method}

\subsection{Multi-Scale Model for Multi-Texture Synthesis}
\label{MTGAN}

Non-stationary textures exhibit various scales of patterns, either among different samples of the same category or in a single sample, such as the peacock feather in Fig.~\ref{fig:intro} (d). It is challenging for single-scale generators~\cite{zhou2018non,Shocher_2019_ICCV} to cope with multiple non-stationary textures. In light of this, we propose to build a generator with the capability to capture multiple scales of structure statistics for non-stationary multi-texture synthesis. 
This is a unique consideration to previous works, partially because they haven't considered the scenario of non-stationary MTS.

% In consideration of the variable scales of structures in different samples, as well as texture blending of different scales in the same sample, single-scale generators~\cite{zhou2018non,Shocher_2019_ICCV} can hardly deal with non-stationary multi-texture synthesis. As a result, a multi-scale generator is proposed with strong scale sensitivity to capture texture features of different scales, named Multi-texture GAN (MTGAN). 

% There exist several typical multi-scale models~\cite{han2008multiscale,heeger1995pyramid} for texture synthesis to generate high-resolution textures or deal with large-scale structures through feature pyramid. Recently, HRNet~\cite{wang2020deep} has demonstrated an efficient multi-branch architecture for various visual recognition tasks.

Our final model, wrapping up this kind of generator, is named as \textbf{M}ulti-\textbf{T}exture \textbf{GAN} (\textbf{MTGAN}). As shown in Fig.~\ref{fig:framework}, it consists of multiple branches (one primary branch and several secondary branches) with different receptive fields.
% Each branch is responsible for modeling a certain scale of structure variance. 
% The output of each branch will be merged together to decode the final expanded texture image.
% Specifically, the proposed MTGAN is depicted in Figure~\ref{fig:framework}.
The primary branch is responsible for image encoding and decoding, while the secondary branches are to provide multi-scale information to the primary branch. Firstly, each branch processes information of different scales via different down-sampling rates. Then, feature maps of different scales are fed into the corresponding residual block~\cite{he2016deep} chains, where each chain consists of six identical residual blocks. 

A critical step is how to fuse the information of different branches in an effective way.
Initially, we try to fuse at the end of the residual block chains. The results are close to the model with only one branch, indicating that the primary branch does not obtain much valid information from secondary branches, as Fig.~\ref{fig:ablation} shown.
% Inspired by the excellent performance of DenseNet~\cite{huang2017densely} which has many connections to bridge information flow, 
To fill up the information gap, we develop a feature fusion strategy to fuse the output of each residual block of different branches in a bottom-up manner. Such fusion can be formulated as
\begin{equation}
F_l= \mathrm{Conv}_{1\times 1}(\mathrm{Concat}(\mathrm{Up}(F_s),F_l)),
\end{equation}
where $F_s$ is the feature map of small scale and $F_l$ is that of the adjacent large scale branch. $F_s$ concatenates with $F_l$ after upsampling, followed by a $1\times 1$ convolution to adjust the number of feature channels. 
With the branch-wise and layer-wise fusion strategy, these branches work collaboratively to make the model sensitive to both local details and global structure. At last, the decoder with an extra upsampling operation transforms the fused feature into an expanded texture image. More details about the architecture can refer to Sec.~\ref{sec:exp:details}.

% \zzhu{Mention alternatives here or in the related work.}
% \noindent \textbf{Difference to some typical multi-scale designs.}
Multi-scale mechanism is a general consideration across many vision tasks. In this work, we consider several options that can be adapted for the non-stationary MTS task.
Some typical examples of multi-scale design for other tasks endorse new network architecture design such as HRNet~\cite{wang2020deep}, insert a multi-scale context module like ASPP~\cite{chen2017deeplab}, {\it {etc}}. 
However, we believe our multi-scale generator is more beneficial to the task. 
% Our MTGAN is architecturally different from the previous works and we had explored some typical methods in Sec.~\ref{chap:ablation}. 
Specifically, each fusion block in MTGAN fuses features of two scales (branch-wise), while HRNet~\cite{wang2020deep} directly fuses features from multiple scales ($>2$ scales) via upsampling in different stages. The upsampling rate may reach up to 8 times, thereby induces much noise and yields apparent artifacts in the results. ASPP~\cite{chen2017deeplab} also performs worse than our method supported by experiments. We hypothesize the common practice of applying such context modules only before the output layer is unable to capture texture scale variance. In contrast, our feature fusion is applied for all ResBlocks (layer-wise) to create enough flexibility for modeling scale variance of non-stationary textures.
The comparison with these methods in the experiment shows the effectiveness of MTGAN.
% Our fusion strategy is also verified to be effective through the ablation experiments.

% It mainly focuses on position sensitive tasks such as semantic segmentation and object detection, while our main concern is global structure and texture details in texture synthesis task. 

% The basic settings are based on~\cite{zhou2018non}, whose ability has been verified in non-stationary single-texture synthesis. 

% We adopt the 6-layers PatchGAN~\cite{isola2017image} discriminator to alternatively train with the generator.
% \zzhu{how much downsampling stages? I'm saying that other people can follow your paper to implement all details. This time please don't get criticized for this since this problem can be easily handled.} 

\subsection{Category-Specific Training}
\label{Training Strategy}

We use three basic loss functions, including adversarial loss~\cite{goodfellow2014generative} $\mathcal{L}_{\mathrm{adv}}$, style loss~\cite{gatys2015texture} $\mathcal{L}_{\mathrm{style}}$, and perceptual loss~\cite{johnson2016perceptual} $\mathcal{L}_{\mathrm{per}}$, as the optimization objectives of our model. All these loss functions are defined between target images and generated images. 
We use adversarial loss to ensure the fidelity of the generated images. The style loss and the perceptual loss are computed upon the representations obtained from a pre-trained VGG~\cite{simonyan2014very} model to secure training stability in early stages. 
The full objective is defined as:
% \vspace{-2ex}
\label{lossfunction}
	\begin{equation}
	\mathcal{L}_{\mathrm{total}}=\arg\underset{G}{\min}\, \underset{D}{\max}\, \mathcal{L}_{\mathrm{adv}} + \mathcal{L}_{\mathrm{style}} + \mathcal{L}_{\mathrm{per}},
	% \mathbb{E}_{x\sim p(x)}
	\end{equation}
where $G$ and $D$ are the generator and the discriminator.

% \zzhu{More details and the beginning of the paragraph seems quite redundant. We should try to shorten it. (Already shortened by me. Maybe you can do the same work later.)}

The existing multi-texture synthesis methods~\cite{li2017diversified,liu2020transposer} randomly select several texture images and then feed them into the neural network for training. However, due to the great difference in the global structure and spatial scale of non-stationary textures from different classes, it is difficult to train a single powerful model that can perform well on all textures. To reduce learning difficulty and improve synthesis quality, we propose a category-specific training strategy. More concretely, only one category of textures is used to train the network during the training phase. In this way, the model can easily grasp the texture structure and features of a specific category by making full use of the complementary information among samples. Moreover, once a model on a specific category is trained, we propose two training strategies to easily extend this model to other instances of the same category or even textures of a different category.
% By default, we conduct experiments under this strategy.

% \vspace{1ex}\noindent\textbf{Adversarial loss.} 

\vspace{1mm}\noindent\textbf{Extension at the instance level from the same category.} Such extension is implemented by a fast fine-tune phase. Specifically, we load the trained model for a category and then input one unseen image of the same class as training data, fine-tuning for several iterations. To avoid catastrophic forgetting, we sample the old data of this category at a frequency of 20$\%$ during the phase. Given this replay-based strategy, not only can our model handle the unseen image, but it can still handle previous samples. Empirically, we found 900 iterations with batch size 4 are enough to make the fine-tuned model support a good expansion of such an unseen image. As shown in Tab.~\ref{tab:time}, our model significantly reduces the time cost on unseen textures compared to STS models.
% Once a model is well-trained on a class of textures, it can be easily extended to arbitrary unseen images of the same category by a fast fine-tune phase. 

\vspace{1mm}\noindent\textbf{Extension at the category level.} 
After we have trained MTGAN on one category of texture images, our approach also allows this trained model to further handle more categories jointly. 
To be more practical and flexible for real application scenarios, the data of old classes are not available but our strategy can still avoid catastrophic forgetting of the old classes when coping with new classes.
Inspired by LwF~\cite{li2017learning} and LwF-GR~\cite{shin2017continual}, we apply a continual learning strategy to achieve this target.
Specifically, we first train a model on a category to a near-optimal state (8000 epochs). Then, we add a new decoder $d^{n}$ to the generator to deal with the coming new category. Note that when the model is trained for the new data, the previous data is not available, thus, we apply a distillation loss $\mathcal{L}_{\mathrm{old}}(Y_o, \hat{Y_o})$ to the old decoder $d^{o}$ to prevent forgetting. $\hat{Y_o}$ is generated by $d^{o}$ when the new model is initialized with old parameters before training, and $Y_o$ is the output from $d^{o}$ during training. $\mathcal{L}_{\mathrm{old}}$ is similar as $\mathcal{L}_{\mathrm{total}}$ besides the $\mathcal{L}_{\mathrm{adv}}$ term and the $\mathcal{L}_{\mathrm{1}}$ regularization term:
\begin{equation}
	\mathcal{L}_{\mathrm{old}}=\arg\underset{G}{\min}\,  \mathcal{L}_{\mathrm{adv}} + \mathcal{L}_{\mathrm{style}} + \mathcal{L}_{\mathrm{per}} + \mathcal{L}_{\mathrm{1}}(D_o, \hat{D_o}).
	% \mathbb{E}_{x\sim p(x)}
	\end{equation}
We compute $\mathcal{L}_{\mathrm{adv}}$ through the old pre-trained discriminator without optimizing the old discriminator since the old discriminator is already knowledgeable to identify generations of the old category. $D_o$ and $\hat{D_o}$ are the outputs by feeding $Y_o$ and $\hat{Y_o}$ to the old discriminator. To deal with the new domain, we let $\mathcal{L}_{\mathrm{new}}=\mathcal{L}_{\mathrm{total}}$. The detailed algorithm of this strategy is depicted in Algorithm~\ref{alg::training}. The continual learning strategy to extend the model to new categories can cater to the emergent open-world scenario~\cite{yang2021generalized}. 

\begin{algorithm}[htb]
      \caption{Continual training strategy}
      \label{alg::training}

      \KwIn{N categories of texture images $X_1$, $X_2$, ..., $X_N$ and the targets $\hat{Y_1}$, $\hat{Y_2}$, ..., $\hat{Y_N}$}
      \textbf{Initialize:} shared parameters $\theta$, old decoder $d^{o}$
      
      Train a generator $G_1(\theta, d^{o})$ on $X_1$

             \For{$n=2;n\leq N$}
             {
                Add a new decoder $d^{n}$
                
                $d^{n} \gets RandInit(|d^{n}|)$ // random initialization
                
                $G_n(\theta, d^{o}, d^{n}) \gets G_{n-1}(\theta, d^{o})$ // load old parameters
                
                Define $\hat{Y_o} \equiv G_{n-1}(X_n, \theta, d^{o})$ // old category target
                
                Define $Y_o \equiv G_{n}(X_n, \theta, d^{o})$ // old category output
                
                Define $Y_n \equiv G_{n}(X_n, \theta, d^{n})$ // new category output
                
                $\theta, d^{o}, d^{n} \gets \underset{\theta, d^{o}, d^{n}}{\arg\min}\,  (\mathcal{L}_{\mathrm{old}}(Y_o, \hat{Y_o}) + \mathcal{L}_{\mathrm{new}}(Y_n, \hat{Y_n}))$
                
                }

\end{algorithm}

\subsection{Multi-Scale Texture Similarity}
\label{chap:3.3}
% It remains a problem that how to evaluate the effect of texture expansion and whether the global structure is preserved in the process of texture synthesis. The mainstream full-reference image quality assessment (IQA) models can be roughly divided into the following categories: pixel-level methods~\cite{wang2009mean}, structural similarity methods~\cite{wang2003multiscale,wang2004image} and deep-feature based methods~\cite{heusel2017gans,ding2020image,zhang2018unreasonable}. MSE~\cite{wang2009mean} calculate the distance among pixel pairs and SSIM~\cite{wang2003multiscale} separate image structure from brightness and contrast, but sometimes they are inconsistent with the subjective evaluation. Deep-feature based methods, ~\emph{eg.} FID~\cite{heusel2017gans} and DISTS~\cite{ding2020image} extract image features via a pre-trained classification network, and then define some operations on them to capture the semantic and structural information. However, as demonstrated in Figure~\ref{fig:user_study}, none of these methods are competent for non-stationary multi-texture expansion.

According to the user study in Fig.~\ref{fig:user_study}, we find currently available metrics for evaluating non-stationary texture expansion performance are not in line with human evaluation. This motivates us to design a more suitable metric for non-stationary expansion evaluation.
Our metric named Multi-Scale Texture Similarity (MSTS) is inspired by FID~\cite{heusel2017gans} to calculate the distance of distributions between two sets of deep features. The first difference is we obtain deep features of the input texture images from a DTD~\cite{cimpoi2014describing}-pretrained DEP model~\cite{xue2018deep} instead of an ImageNet~\cite{deng2009imagenet}-pretrained Inception~\cite{szegedy2016rethinking} model. Texture images are not object-centric as ImageNet images, thus the DTD-pretrained model is more related to our task.
% Considering the domain gap, we pre-train a classification model~\cite{xue2018deep} utilizing a challenging dataset DTD~\cite{cimpoi2014describing}. 

% Instead of Inception, we draw lessons from a meaningful texture recognition model to obtain robust texture feature representation, which captures texture and spatial features simultaneously.

% Specifically, DEP model consists of three parts: encoding layer, average pooling and bilinear model. The encoding layer encodes texture details by dictionary learning. The average pooling layer map the visual descriptors into a spatial feature via a $7\times7$ convolution kernel. Descriptors from convolutional layers are fed into the encoding layer and the average pooling layer jointly. Then, the output texture information and spatial information are combined through a bilinear model. Notice that the size of the spatial feature is determined by that of the input image, so the size of feature used for texture synthesis evaluation is also variable. In this work, the number of feature channels is set to 128 by default. 

% referring to the statements in the papers of InGAN~\cite{Shocher_2019_ICCV} and SinGAN~\cite{shaham2019singan}
Another major difference is FID calculates the distance of distributions between two sets of \emph{images} while MSTS calculates the similarity between the internal distributions of \emph{patches} within the original image and the generated image.
% Unlike FID to calculate the distance of distributions between two sets of images, our main idea for MSTS is to calculate the similarity between the internal distributions of patches within the original image and the generated image. 
Given the internal distributions of the original texture and the generated texture as $p_o(\cdot)$ and $p_g(\cdot)$, respectively. The equality $p_o(\cdot)=p_g(\cdot)$ holds \emph{iff.} $\int p_o(\cdot)f(x)dx=\int p_g(\cdot)f(x)dx$ where $f(x)$ is arithmetic operations of input data $x$. 
%\zzhu{What do you want to show here? what is $x$? You mean for the distributions of both $p_o$ and $p_g$ to be equal, it is must be ensured that the distributions of their patches should also match? I do not like the current form. Try to use more formal deduction and be clear of your goal.}
Here we replace $x$ with the 128-dimensional feature vector output from the penultimate layer of the DEP model. We assume the feature to follow a multidimensional Gaussian, and we utilize Wasserstein-2 distance to measure the similarity between these two distributions. Our final metric is thus given by
\begin{equation}
s= (\|m_o-m_g\|^2_2 + \mathrm{Tr}(C_o+C_g-2(C_oC_g)^{1/2}))^{1/2},
\end{equation}
where $m_o$ and $C_o$ are the mean and covariance of $p_o(\cdot)$. Analogously, $m_g$ and $C_g$ are the mean and covariance of $p_g(\cdot)$.

Furthermore, multi-scale statistics are helpful for capturing structure and scale variance~\cite{wang2003multiscale}. To this end, MSTS is designed with a multi-scale mechanism, which is also hugely different from FID. In addition to calculating scores on the full image, we also randomly crop patches of different scales with 16 samples for each scale, where the scales are set to $\frac{1}{4}$, $\frac{1}{2}$, $\frac{3}{4}$ of the original image. Finally, we have four scores $s_1$, $s_2$, $s_3$, $s_4$ corresponding to four different scales ranging from large to small. MSTS score can be expressed as a combination of the four scores:
\begin{equation}
\mathrm{MSTS} = (s_1+s_2+s_3+s_4)\times 10^3.
\end{equation}
Note we multiply MSTS by $10^3$ to enlarge the final value to make it more intuitive.
Such a metric can jointly evaluate global texture structure and local texture details as verified through our user study.

%where $r$ is a scaling factor. When we apply MSTS to evaluate the global structure, the overall layout of textures should be paid more attention to. Thus, $0<r<1$ is our choice and we set $r=0.8$ by default. On the other hand, we need to take care of more local details when we evaluate whether the texture is effectively expanded or simply zoomed in as the global structure is usually not destroyed in this case. Consequently, $r>1$ is more appropriate and $r=1.2$ is set by default.

\section{Experiments}

\subsection{Datasets and Implementation Details}
\label{sec:exp:details}
\begin{figure}[t]
	% \vspace{-0.5cm}
	\centering
	\includegraphics[width=0.49\textwidth]{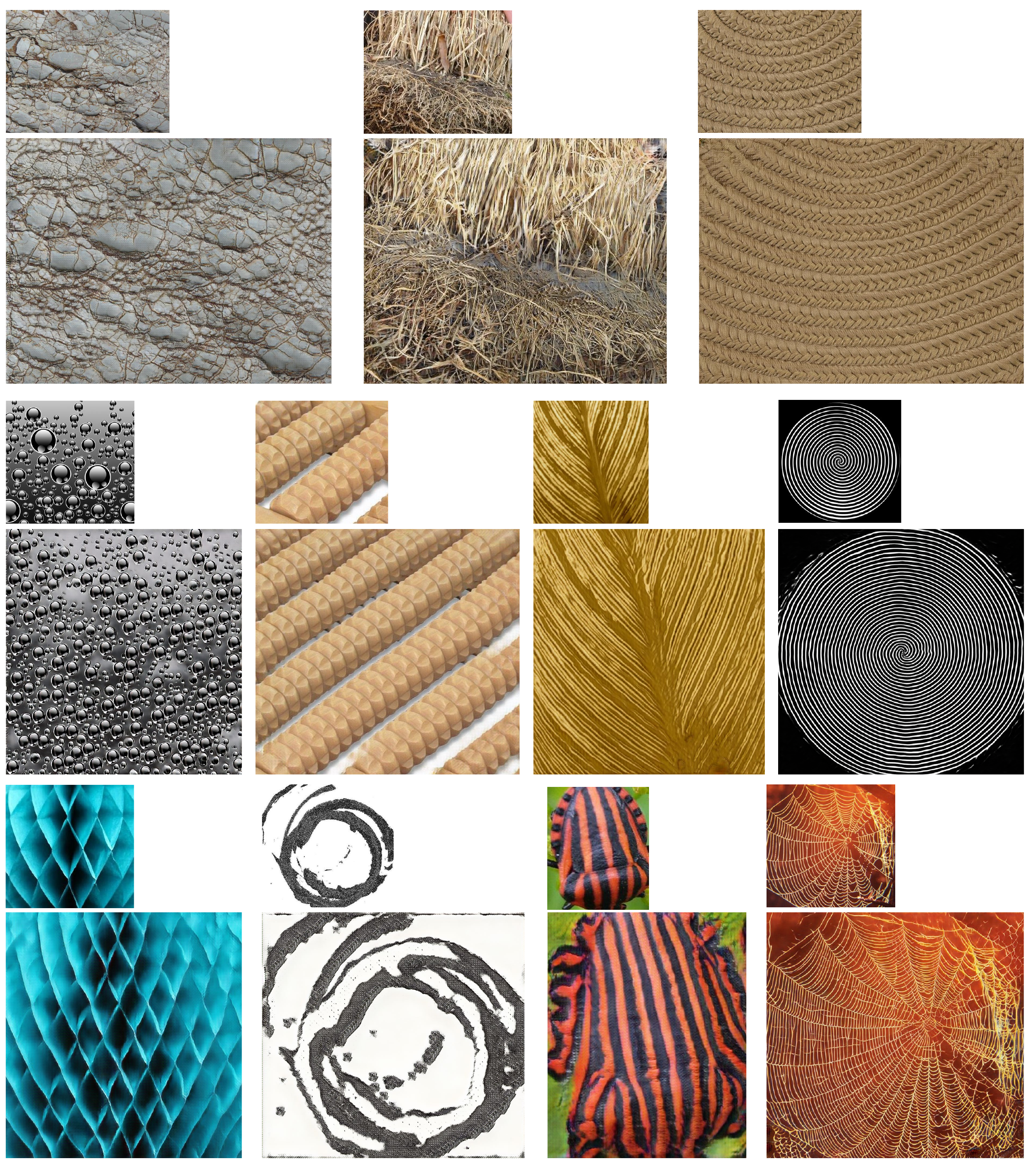}
	\caption{Texture expansion results presentation on {\bf DTD-11}}
    \label{fig:more_results}
\end{figure}

Our experiments are conducted on Describable Texture Dataset (DTD)~\cite{cimpoi2014describing}, which consists of 47 texture categories with 120 images in each class. This dataset contains a large number of non-stationary texture images with complex backgrounds and various scales. Considering the feasibility of the experiments for evaluating non-stationary multi-texture synthesis performance, we form a subset by choosing 11 representative and challenging non-stationary classes: braided, bubbly, bumpy, cobwebbed, cracked, fibrous, honeycombed, scaly, spiralled, stained, veined. We denote this selected dataset as {\bf DTD-11}. Fig.~\ref{fig:more_results} shows the results of our approach on {\bf DTD-11} dataset. The proposed method can cope with various challenging non-stationary textures.
Besides, we also collected some challenging texture images from the Internet to test the generalization ability of our method on unseen images. 

\begin{figure}[t]
	% \vspace{-0.5cm}
	\centering
	\includegraphics[width=0.48\textwidth]{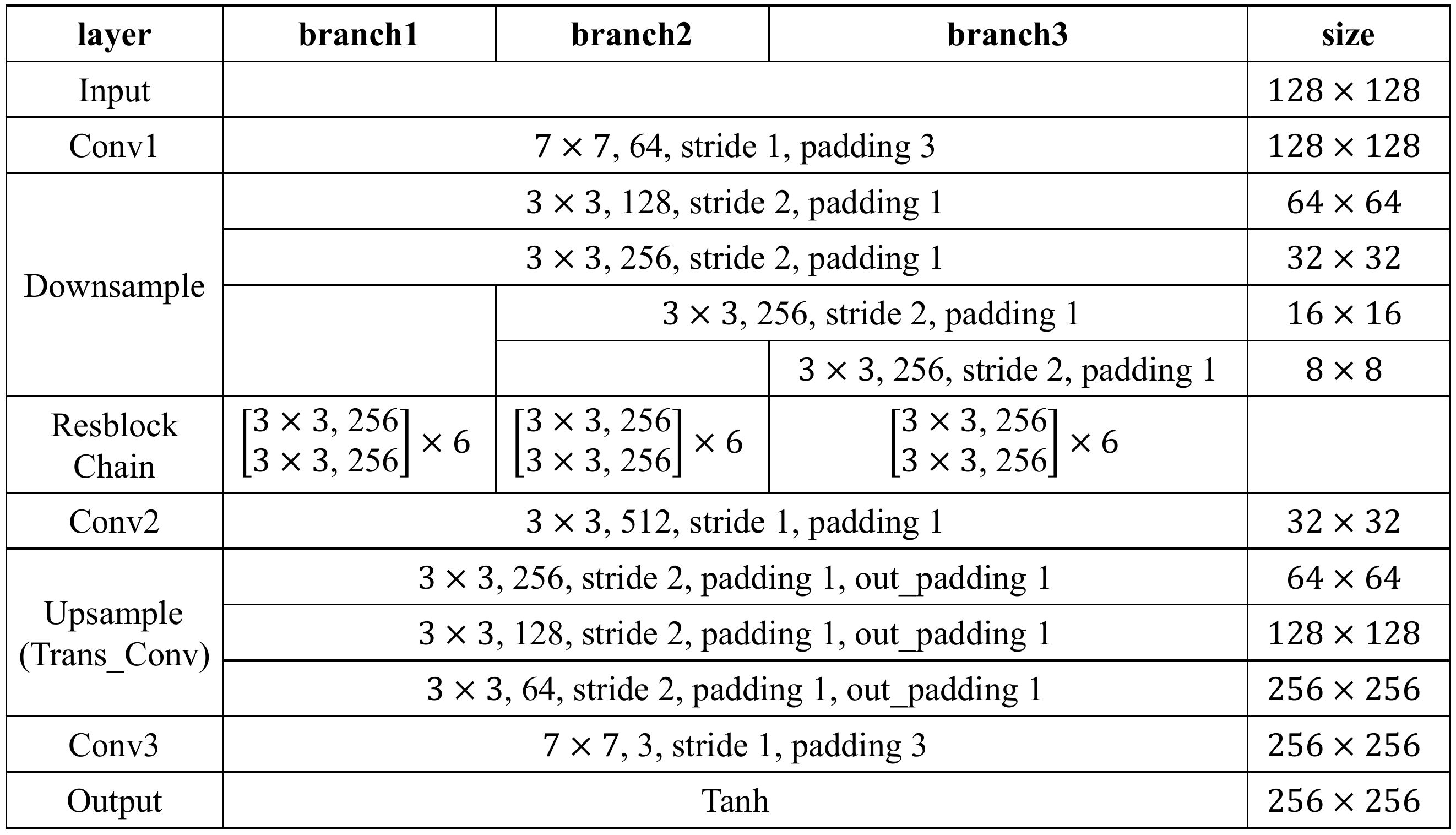}
	\caption{Architecture details of our multi-scale generator.}
    \label{fig:architecture}
\end{figure}

The detailed generator architecture of MTGAN is shown in Fig.~\ref{fig:architecture}. We adopt the 6-layers PatchGAN~\cite{isola2017image} discriminator to alternatively train with the generator.
We train our model using Adam~\cite{kingma2014adam} optimizer, where its hyper-parameters are set as $lr=0.0002$, $\beta_1=0.5$ and $\beta_2=0.999$. We train each category of textures individually for 8000 epochs with batch size 4 on one NVIDIA Titan Xp. The learning rate keeps unchanged for the first 4000 
% \zzhu{iteration or epoch? should be clear!} 
epochs and then linearly decays to zero over the remaining 4000 epochs. 

% Unlike the common practice, we use instance normalization rather than batch normalization because the latter causes color distortion when training multiple textures.
% input mini-batches to the model instead of a single sample.

\subsection{Evaluation of MSTS}

\begin{figure}
	% \vspace{-0.5cm}
	\centering
	\includegraphics[width=0.48\textwidth]{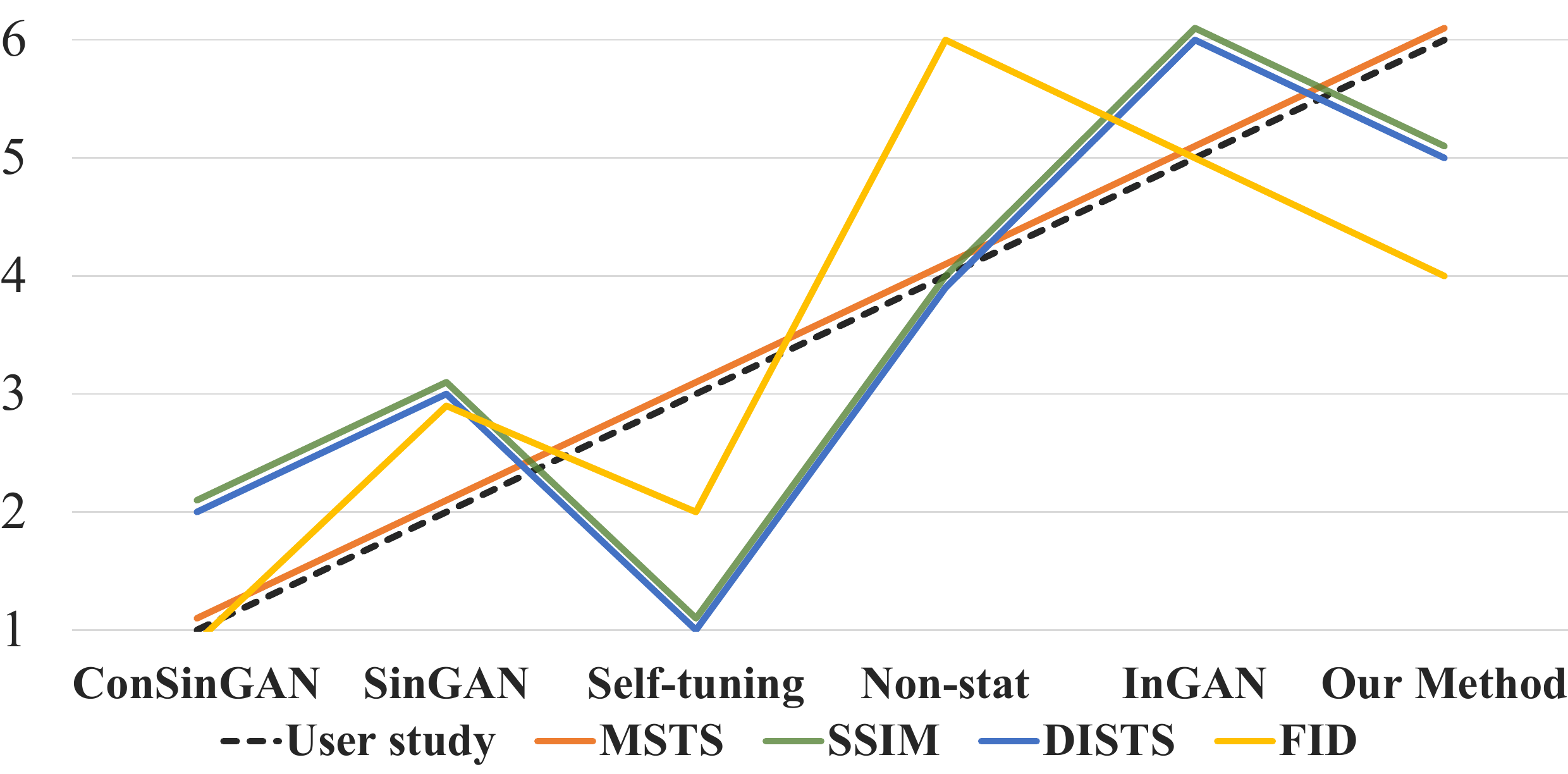}
	\caption{Different quantitative metrics and user study. The $x$-axis represents six different models, and the $y$-axis represents the ranking of the models under different metrics.}
    \label{fig:user_study}
\end{figure}

We evaluate the effectiveness of MSTS for texture synthesis tasks via a user study. We first collect 5 images of each category generated by each of the following models: self-tuning~\cite{kaspar2015self}, Non-stat.~\cite{zhou2018non}, InGAN~\cite{Shocher_2019_ICCV}, SinGAN~\cite{shaham2019singan}, ConSinGAN~\cite{hinz2020improved}, and our model. We then create a questionnaire with 55 questions, where each question asks how well the generated image is expanded according to the original structure and texture arrangement. Then, we distribute the questionnaire to twenty users to sort the images of each question from 1 (least pleasant) to 6 (most satisfactory). Tab.~\ref{tab:single} (Pref.) shows the average preference scores for the user study. 
% \zzhu{The following sentence is a bit unclear. Please modify it.}
Besides, we also evaluate the performance of these models on several objective metrics: SSIM~\cite{wang2003multiscale}, FID~\cite{heusel2017gans}, DISTS~\cite{ding2020image} and MSTS. To verify the effectiveness of MSTS, we rank all the models according to each metric, as shown in Fig.~\ref{fig:user_study}. The sorting result of MSTS is more consistent with our user study, which demonstrates MSTS is comparatively more suitable for evaluating non-stationary expansion results.

\begin{figure}
	% \vspace{-0.5cm}
	\centering
	\includegraphics[width=0.48\textwidth]{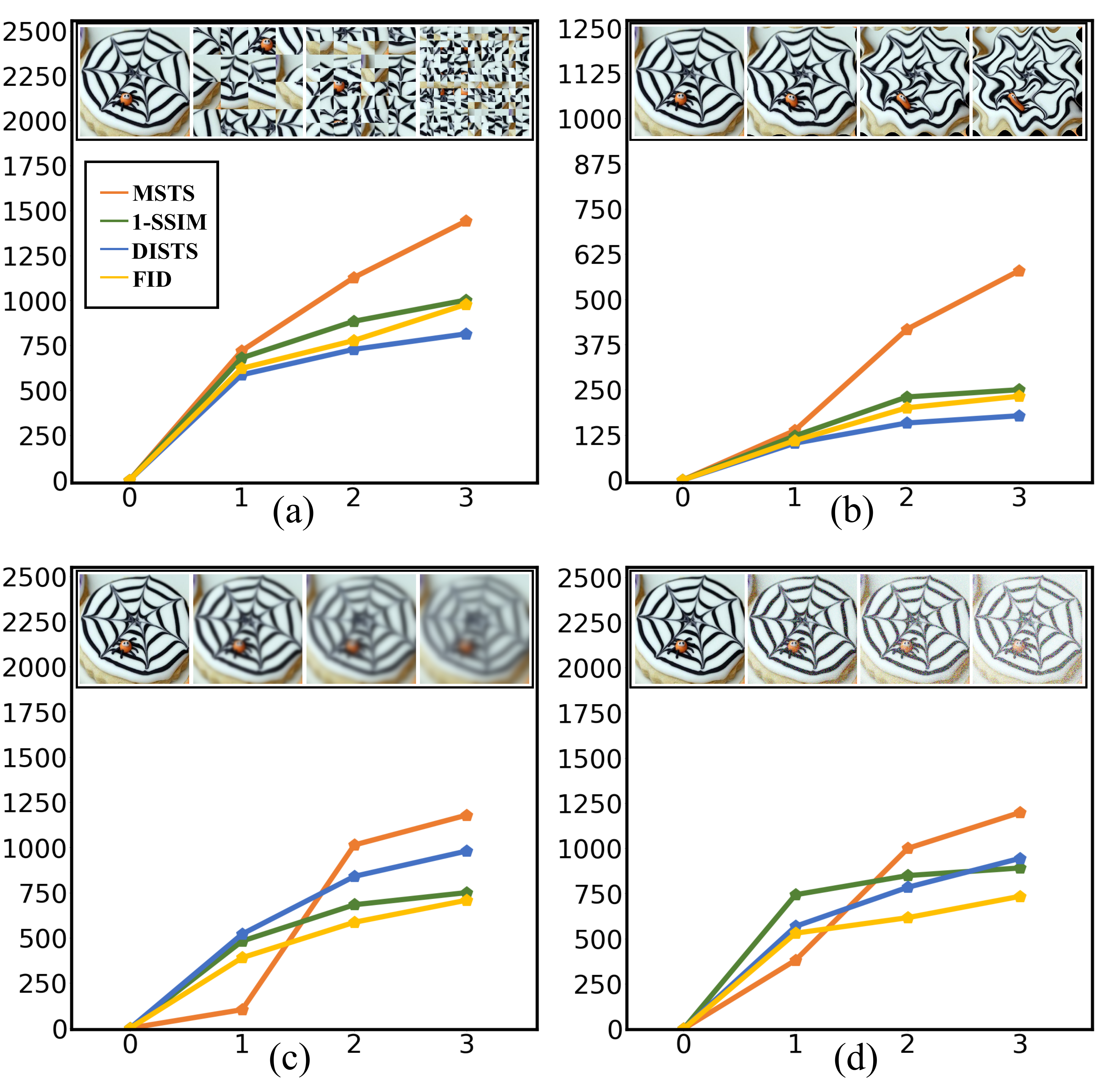}
	\caption{MSTS is evaluated for (a): Shuffling image patches, (b): Image warping, (c): Gaussian blurring, (d): Removing pixels. The disturbance level linearly rises from zero and increases to the highest level. All the metrics are linearly scaled to the same magnitude.}
    \label{fig:metrics}
\end{figure}

%Finally, comparing the user study results with those given by metrics including SSIM~\cite{wang2003multiscale}, FID~\cite{heusel2017gans}, DISTS~\cite{ding2020image}, their conclusions are not consistent, as shown in Figure~\ref{fig:user_study}. On the contrary, the proposed MSTS is more consistent with user preferences, which demonstrates MSTS is comparatively more suitable for evaluating non-stationary expansion results.

% in evaluat the global structure and local quality of the results more robustly, and is more consistent with user study tendency. MSTS has more relevant results with user study than other metrics. 

In addition, we test the sensitivity of MSTS and other three metrics when encountering textures with structure or texture distortions. To test \emph{global structure sensitivity}, we apply two distortion operations: shuffling texture patches and image warping.
We also apply two other distortions to test \emph{texture change sensitivity}: Gaussian blurring and randomly removing pixels.
% To test the global structure sensitivity of all metrics, we randomly shuffle image patches of different sizes. 
As shown in Fig.~\ref{fig:metrics}~(a) \& (b), when the distortion is weak (disturbance level~\cite{heusel2017gans} is 1), MSTS value is competitive to other metrics. When the structure distortion becomes more severe, MSTS makes the highest responses. Also, note that the curve slopes of other metrics after disturbance level 1 are quite smaller than that of MSTS. These two plots unveil that MSTS is more sensitive to large structure distortion than other metrics.
% As for Figure~\ref{fig:metrics}(b), when the disturbance level is 1, the structure change is minor
% The more chaotic the patches, the worse the global structure. 
% \zzhu{Starting here, describe the texture distortion.}
Moreover, MSTS presents more reasonable responses to texture distortions, as show in Fig.~\ref{fig:metrics}~(c) \& (d). When the disturbance level is 1, the subjective visual difference of the images is subtle, resulting in a smaller value of MSTS. By increasing the disturbance level, the image quality decreases significantly, and MSTS shows the most sensitive responses. Consequently, MSTS can be a more suitable and effective metric for the texture synthesis task.

Since MSTS estimates the internal patch distribution within a single image, 16 patch samples for each spatial scale can already obtain stable results. We conduct MSTS evaluation of ``Ours (Multi)'' in Tab.~\ref{tab:single} w.r.t. different sample sizes with 50 random seeds for each sample size. Then we calculate the mean and standard deviation for each sample size over 50 seeds (size@mean/std): 8@99.46/4.50, 16@98.34/2.13, 32@98.28/1.80, 64@98.51/1.17. Actually, the results reported in the paper are the mean values over 50 random seeds. We can obtain more stable results by increasing the sample size but the evaluation time will increase.

\subsection{Evaluation of MTGAN}

\begin{figure*}[t]
	% \vspace{-0.5cm}
	\centering
	\includegraphics[width=0.95\textwidth]{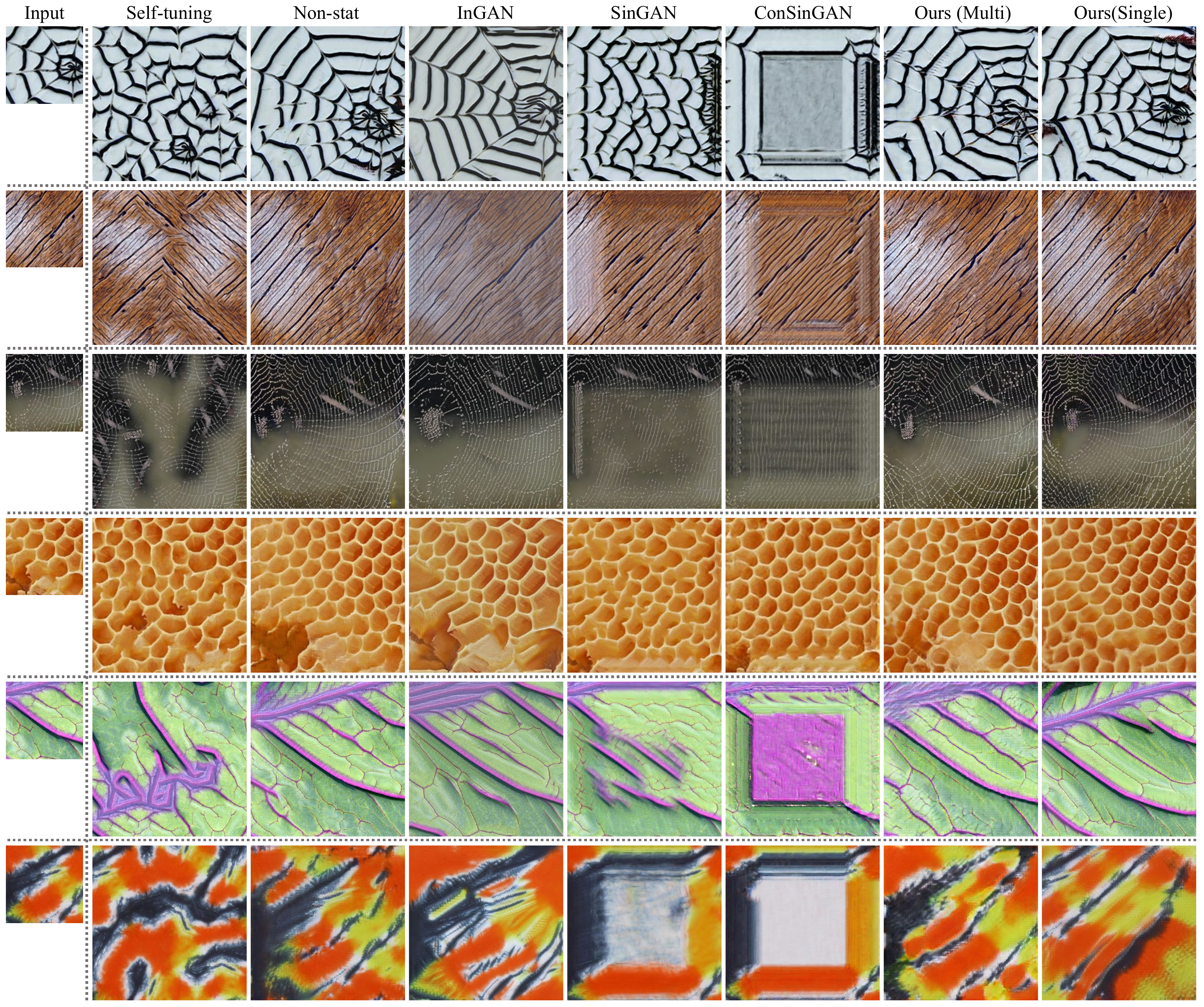}
% 	\vspace{-4ex}
	\caption{Qualitative comparisons with single-texture synthesis methods. Our approach can achieve better expansion results than the STS models on non-stationary textures and substantially improve the time efficiency when training and extending to unseen image.}
    \label{fig:single_comparison}
    % \vspace{-1ex}
\end{figure*}

\begin{table}
    \centering
    % \footnotesize
    \small
    % \resizebox{10pt}{50mm}{
    \setlength\tabcolsep{2.2pt}
    \caption{Quantitative results of STS methods. Hist. means Color Histogram comparison between the synthesis images and inputs. Pref. means preference scores from the user study.}
    \begin{tabular}{|l|c|c|c|c|c|c|}
    \hline
        Methods & MSTS $\downarrow$ & SSIM $\uparrow$ & DISTS $\downarrow$  & FID $\downarrow$  & Hist. $\uparrow$ & Pref. $\uparrow$ \\ \hline
        Self-tuning & 109.43 & 0.0539 & 0.3653 & 288.7 & {\bf 0.975} & 2.20 \\ 
        Non-stat. & 107.99 & 0.0725 & 0.3156   & 209.6  & 0.827 & 2.85 \\
        InGAN & 103.48 & {\bf 0.1818} & 0.3074  & 222.3 & 0.667 & 2.91\\
        SinGAN & 126.83 & 0.0664 & 0.3465  & 259.1 & 0.844 & 2.08\\
        ConSinGAN & 205.14 & 0.0561 & 0.3649  & 337.2 & 0.860 & 1.91\\  \hline
        Ours (Multi) & 98.34 & 0.0831 & 0.3141  & 233.4 & 0.838 & {\bf 3.55}\\
        Ours (Single) & {\bf 97.68} & 0.0845 & {\bf 0.3061}  & {\bf 196.9} & 0.862 & - \\ \hline
    \end{tabular}
    % }
    % \vspace{0.05cm}
    % \normalsize
    
    \label{tab:single}

\end{table}

% \zzhu{The numerical values in Tab. 1 should be more convincing.}
% \subsubsection{Comparison with Single-Texture Synthesis Model }
% In this part, we train 10 models separately on each class of DTD-10, to compare our method with single-texture and multi-texture synthesis methods qualitatively and quantitatively. We evaluate these models by inputting cropped $256\times 256$ images to generate outputs with a size of $512\times 512$. We use MSTS, SSIM, DISTS, and FID to comprehensively evaluate the image quality. 
We compare our MTGAN with STS and MTS methods qualitatively and quantitatively. We use MSTS, SSIM, DISTS, and FID to evaluate the quality of the synthesized textures comprehensively. Besides, we also compare the time efficiency of these methods in the training and inference phases. These comparisons show the superiority and practicability of our method on non-stationary texture synthesis.

% \zzhu{should highlight here our results on single texture are competitive to other SOTAs, yet our method support multi-texture synthesis. For multi-texture comparison, we should mention that our method supports non-stationary textures. Also mention more training and testing settings.}

%  \vspace{-0.2cm}

\vspace{1mm}\noindent\textbf{Compare with single-texture synthesis methods.}~Since STS methods can usually deal with non-stationary textures and generate better results than MTS methods, we compare MTGAN with them to demonstrate our high generation quality and superior time efficiency.
We follow the evaluation protocol of \cite{liu2020transposer}, also an MTS method to compare with STS methods. Besides the metrics mentioned above, we utilize color histogram comparison and user study to comprehensively assess these methods. 
The recent related competitors for STS are self-tuning~\cite{kaspar2015self}, Non-stat.~\cite{zhou2018non}, InGAN~\cite{Shocher_2019_ICCV}, SinGAN~\cite{shaham2019singan} and ConSinGAN~\cite{hinz2020improved}. 
We follow the implementation of these methods in their papers. Considering the fact that STS models are too slow to be trained on all images of DTD-11, we randomly select 55 samples uniformly from 11 categories to evaluate. 
To compare with them, we train our model under two settings: 1) \textbf{Multi}: model trained on all the data of each category in DTD-11 separately. We then select the 55 corresponding images from the synthesis results for comparison. Note, our model is not fine-tuned for these images. 2) \textbf{Single}: model trained on a specific image for 10000 iterations by completely analogy to the training strategy of other STS models. 
% Note that our models are trained on DTD-10 to expand all the images with generalization ability, while other methods are trained on a single image to overfit the model so as to manipulate a specific image~\cite{shaham2019singan}.

% \zzhu{This part may leave readers wonder what about then training MTGAN on these 50 samples just like what you do to single image methods? Though overfitting is quite easy for CNN, you actually used more data, which makes the comparison not as fair as you claimed. You should also tune down a little bit. I'm not sure everyone has the same feeling that generalization is more difficult than overfitting. }
For inference, we randomly crop a $256\times 256$ patch for each image and feed it into these models to generate outputs with a size of $512\times 512$.
As Fig.~\ref{fig:single_comparison} shows, our model performs well in various non-stationary textures with different scales and categories. While preserving the global structure of the input textures, our results exhibit fewer artifacts. Self-tuning 
% can produce textures with good visual quality, but the generated images 
cannot follow the global structure, and its results tend to be highly repetitive. SinGAN and ConSinGAN are incompetent for textures with large-scale structure, and their results retain only basic lines and color information. InGAN maintains the global structure, but it suffers from obvious color deviation, for instance, the first row of InGAN is darker than the input and the second row is brighter. Non-stat. achieves balanced results, but some show unreasonable patterns. For example, the cobwebbed images in the first and third rows present two centers while it is known that cobweb generally has only one center, and there are many damaged honeycombed parts in the lower right corner of the fourth row. Together, the results establish the superiority of our multi-scale design for MTGAN because the model can maintain the layouts and color distribution of original textures, at the same time, performing natural expansion.
% This further demonstrates the superiority of training on multiple images of each class because the model may gather useful information about the texture, such as the typical layouts, color distribution, and how to expand. 
% \zzhu{should mention single vs. multi. Also, you should place single at the left hand of Multi...since multi is your method.}

We quantitatively evaluate all the methods on these 55 images, and show the average scores in Tab.~\ref{tab:single}. Our method achieves SOTA performance. 
% Noting that all the SSIM values are small because SSIM is used to evaluate the similarity of an image pair with the same content and different distortions. 
SSIM is computed by resizing the generated images to their input sizes. As a result, it has small reference value since a result that is simply the super-resolution of the input would obtain the highest score.
Compared with MTGAN (Multi), though InGAN performs better in DISTS, FID, it presents severe color distortion problem as hinted by the Hist. in Tab.~\ref{tab:single} and thus performs worse than our method in the user study. The phenomenon is also observed in Fig.~\ref{fig:single_comparison}. Besides, our approach achieves significantly better time efficiency than the STS models when training and extending to unseen images, as shown in Tab.~\ref{tab:time}.

% \textcolor{blue}{Questionable here. Consider deleting the following sentence. Will only make me wonder what is the performance of modifying these approaches for MTS.} Also note, InGAN and other STS methods haven't tested for multi-texture synthesis, which is the major drawback as compared to our approach.

% and they are obviously inferior to our method in time efficiency (Table~\ref{tab:time}). 
% \zzhu{This is not a direct cause. Do not write it this way. You should mention at the beginning of the comparison that you include color histogram and user preference as the evaluation.} 

% \zzhu{After reading the whole section, I would wonder what if these methods are wrapped with category-wise training? Could they do even better? Actually this is the correct way to do the experiment.}

% \vspace{1ex}\noindent\textbf{Quantitative Comparisons.}

% \vspace{1ex}\noindent\textbf{Qualitative Comparisons.}

% \subsubsection{Comparison with Multi-Texture Synthesis Model }
\begin{figure}
	% \vspace{-0.5cm}
	\centering
	\includegraphics[width=0.48\textwidth]{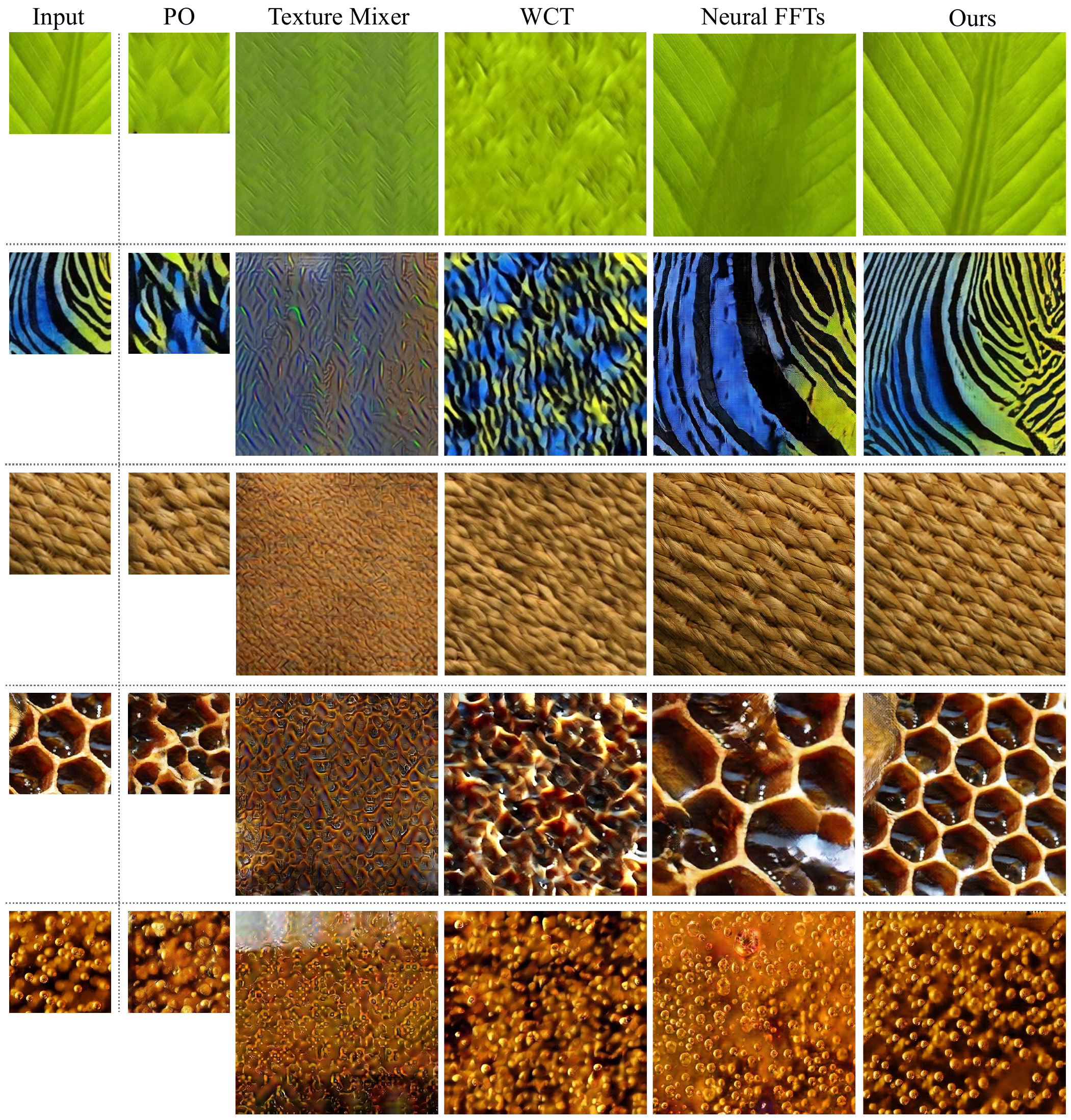}
% 	\vspace{-1ex}
	\caption{Qualitative comparisons with multi-texture synthesis methods. }
    \label{fig:multi-comparison}
    % \vspace{-1ex}
\end{figure}

\begin{table}
    \centering
    \small
    \setlength\tabcolsep{5pt}
    \caption{Quantitative results of MTS methods.}
    \begin{tabular}{|l|c|c|c|c|}
    \hline
        Methods & MSTS $\downarrow$ & SSIM $\uparrow$ & DISTS $\downarrow$  & FID $\downarrow$  \\ \hline
        PO & 345.13 & 0.0431 & 0.3610 & 181.58 \\    
     Texture Mixer & 319.28 & 0.0494 & 0.4448  & 231.18 \\ 
        WCT & 223.39 & 0.0748 & 0.3869  & 219.18 \\  
        Neural FFTs & 206.83 & 0.0692 & 0.3758 & 164.84 \\ \hline
        Our method & {\bf 156.66} & {\bf 0.0911} & {\bf 0.3323}  & {\bf 128.93} \\ \hline
    \end{tabular}
    % \vspace{0.2cm}
    % \normalsize
    % \vspace{-1ex}
    \label{tab:multi}
    % \vspace{-1ex}
\end{table}

\begin{table}
    \centering
    \small
    \setlength\tabcolsep{5pt}
    \caption{Time efficiency comparisons. Unseen img. means the time for extending the model to a new image. {\bf Note:} each value shows the average time on one image.}
    \begin{tabular}{|l|c|c|c|}
    \hline
        Methods & Training & Inference  & Unseen img.   \\ \hline
        Self-tuning & - &  \hspace{1.5mm} 4.5 \hspace{0.5mm} mins \hspace{1mm} & \hspace{1.5mm} 4.5 \hspace{0.5mm} mins \hspace{1mm} \\ 
        Non-stat. & 5.0 \hspace{1mm} hrs & 51.6 \hspace{1mm} ms & 5.0 \hspace{1mm} hrs \\
        InGAN & 4.1 \hspace{1mm} hrs &  430 \hspace{1mm} ms & 4.1 \hspace{1mm} hrs \\
        SinGAN & 2.5 \hspace{1mm} hrs &  379 \hspace{1mm} ms & 2.5 \hspace{1mm} hrs \\
        ConSinGAN &   0.9 \hspace{1mm} hrs &  200 \hspace{1mm} ms & 0.9 \hspace{1mm} hrs \\  \hline
        PO & 0.42 \hspace{0.5mm}  hrs & 92 \hspace{2mm} ms & 92 \hspace{2mm} ms \\ 
        Texture Mixer & 1.25 \hspace{0.5mm}  hrs  & 850 \hspace{1mm} ms & - \\ 
        WCT & - & 8 \hspace{2mm} s & 8 \hspace{3mm} s \hspace{1mm} \\  
        Neural FFTs & 0.97 \hspace{0.5mm}  hrs & 210 \hspace{1mm} ms & 210 \hspace{1mm} ms \\ \hline
        Our method & 0.18\hspace{1mm}  hrs & 60 \hspace{1mm} ms & \hspace{1mm} 5 \hspace{2mm} mins \\ \hline
    \end{tabular} 
    % \vspace{0.2cm}
    % \vspace{-1ex}
        
    \label{tab:time}
\end{table}

\footnotetext[1]{Since the official code of Neural FFTs~\cite{mardani2020neural} is not available, we reproduced its model by ourselves and report the results.}
\vspace{1mm}\noindent\textbf{Compare with multi-texture synthesis methods.} 
MTS methods can handle multiple samples within one model. 
%As a kind of these, our method constrains this task to a single domain. 
To demonstrate the powerful ability of our MTGAN to cope with non-stationary textures, we compare MTGAN with some recent and typical MTS methods such as WCT~\cite{li2017universal}, Texture Mixer~\cite{yu2019texture}, PO~\cite{shi2020fast}, and Neural FFTs~\cite{mardani2020neural}\footnotemark[1]. For a fair comparison, all the models are optimized through the proposed category-specific training strategy. Other settings of training details follow their original papers. PO cannot be applied to texture expansion and its outputs are resized to the sizes of inputs.

Tab.~\ref{tab:multi} indicates that our MTGAN outperforms other MTS methods under all metrics. The qualitative comparisons in Fig.~\ref{fig:multi-comparison} show that all the other methods cannot deal with {\it non-stationary} textures, without preserving the global structure or just enlarging the input images. Texture Mixer mainly retains the color information and few texture details of the input texture. The structure is seriously destroyed, and only stationary textures can be generated. Though better than Texture Mixer, WCT shares the same problem. According to the observation in Fig.~\ref{fig:multi-comparison} and the statement in the original paper~\cite{mardani2020neural}, Neural FFTs cannot expand non-stationary textures and produces many artifacts. In contrast, our method synthesizes natural expansions without destroying global structures and producing artifacts. Also, the time efficiency of our method achieves state-of-the-art performance compared to other MTS methods.
% as shown in Tab.~\ref{tab:time}.

% Furthermore, the training and inference time of MTGAN is more advantageous to practical applications.

%Qualitative comparisons results are shown in Fig.~\ref{fig:multi-comparison}. As can be seen, all the other methods cannot preserve the pattern structure. PO only can generate images have the same size as the input, so it can't be applied to texture expanding. At the same time, it uses the gram matrix to guide image synthesis which is more utilized in style transfer task and can't preserve the structure. Since the Texture Mixer applies the shuffling procedure, it will break the structure of the image. Also, the whitening and coloring transforms proposed at WCT cannot preserve the sample image structure. In contrast, our method can not only generate more realistic images but also preserve the input structure. 

\vspace{1mm}\noindent\textbf{Time efficiency comparisons.} We count the training and inference time of all the STS and MTS models in Tab.~\ref{tab:time}. For STS models, we evaluate the training time on a single image. For MTS models, we get the training time on a category of textures (120 images) and then compute the average time on one image. The inference time is calculated by generating an image of size $512\times 512$. All the evaluations are conducted on one NVIDIA Titan Xp with an Intel Xeon CPU E5-2637 @ 3.50GHz. Compared with STS methods, MTGAN reduces the training time of each texture by 5-27 times. Meanwhile, our inference time is competitive to that of Non-stat. and outperforms others, indicating that our multi-scale design does not harm inference speed.
Compared to MTS methods, MTGAN shows obvious superiority both in training and inference time, which reveals that it is better and more applicable to real-world scenarios.

\subsection{Evaluation of training strategies}
\begin{figure*}
	% \vspace{-0.5cm}
	\centering
	\includegraphics[width=0.95\textwidth]{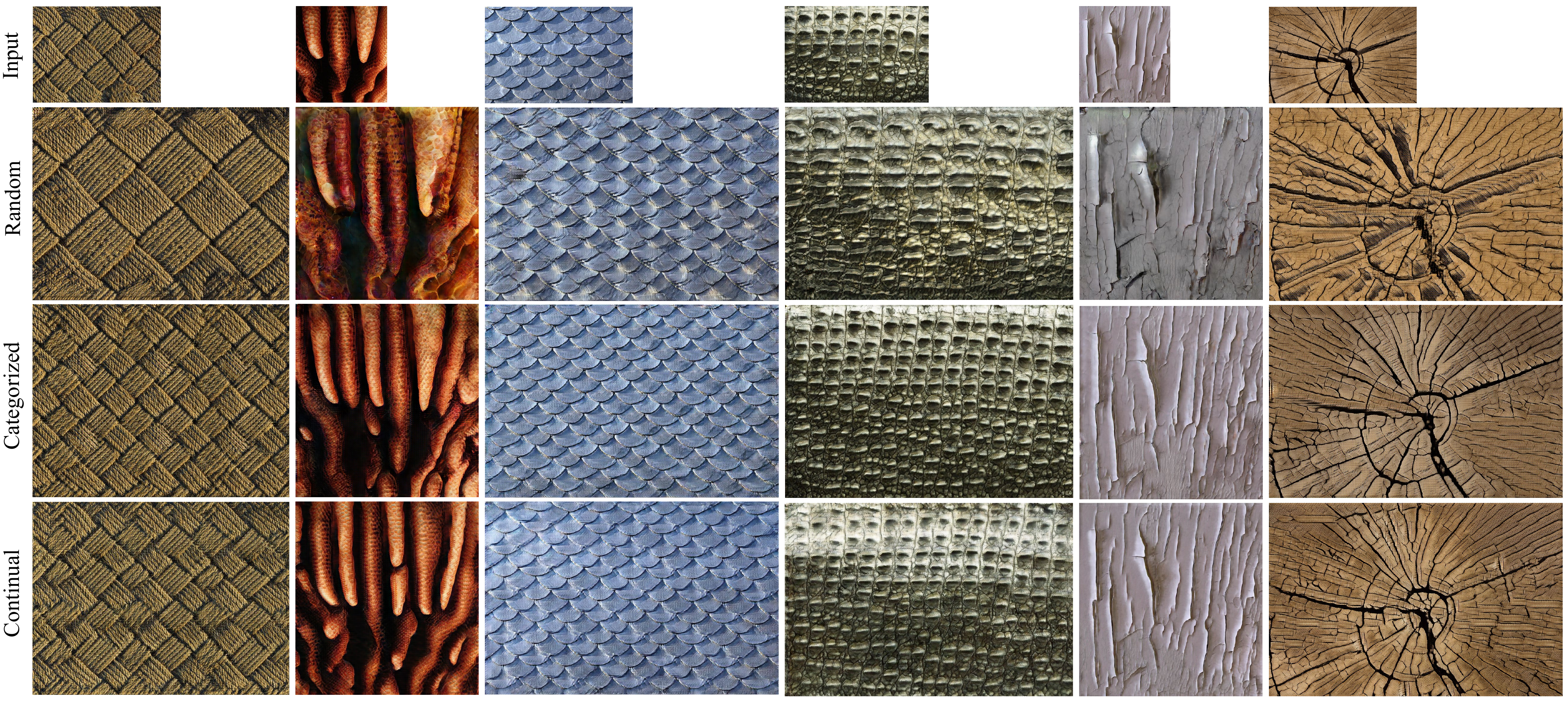}
% 	\vspace{-3ex}
	\caption{Results of different training strategies. The results obtained by continual learning show satisfactory effects, though with a slight decrease compared to the category-specific training strategy, and both the training strategies are significantly superior to random training.}
    \label{fig:train_strategy}
    % \vspace{-1ex}
\end{figure*}
This part is to show the effectiveness of the category-specific training strategy, the fast fine-tune strategy for instances in the single category, and the continual learning strategy to extend a trained model to a different domain.

% To verify the effectiveness of the category-specific training strategy, we conduct an experiment to compare with the random training strategy. 
\vspace{1mm}\noindent\textbf{Effectiveness of the category-specific training strategy.} One straightforward strategy to optimize our model is \emph{random training}, where we randomly select 110 images from the whole DTD-11 and mix them together to train MTGAN; For \emph{category-specific training}, the 110 training images are from the same category. Fig.~\ref{fig:train_strategy} shows several results of different training strategies. Compared with random training, category-specific training achieves apparently better results.
% For example, the center of the cobweb is not well expanded by random training.

% \zzhu{Not clear of the goal of this experiment. SHould be organized or merged into other sections if needed. }
% Our approach also allows a model to handle multiple categories jointly. We apply a simple replay-based incremental training strategy~\cite{rebuffi2017icarl,li2017diversified} to achieve this target. Specifically, we first train a model on a category to a near-optimal state. Then, we gradually feed data from new categories to the model without letting it forget the previous categories in the meantime. Every time adding a new category, we will sample data from previous categories with a uniform probability. We always ensure that the sampling probability of the new category is $\frac{1}{2}$, and the rest $\frac{1}{2}$ is assigned to all old categories with equal sampling probability. The detailed algorithm of this incremental training strategy is illustrated in supplementary materials.

% Once a model is trained on a class of textures, it can be easily extended to arbitrary unseen images of the same category by a fast fine-tune phase. Specifically, we first load the pre-trained model, and then input one unseen image as training data to fine-tune for several iterations. To avoid catastrophic forgetting, we also sample the old data of this category at a frequency of 20$\%$ during the phase. Empirically, we found 900 iterations are enough to make the fine-tuned model support a good expansion of such an unseen image. 

\begin{figure}
	% \vspace{-0.5cm}
	\centering
	\includegraphics[width=0.48\textwidth]{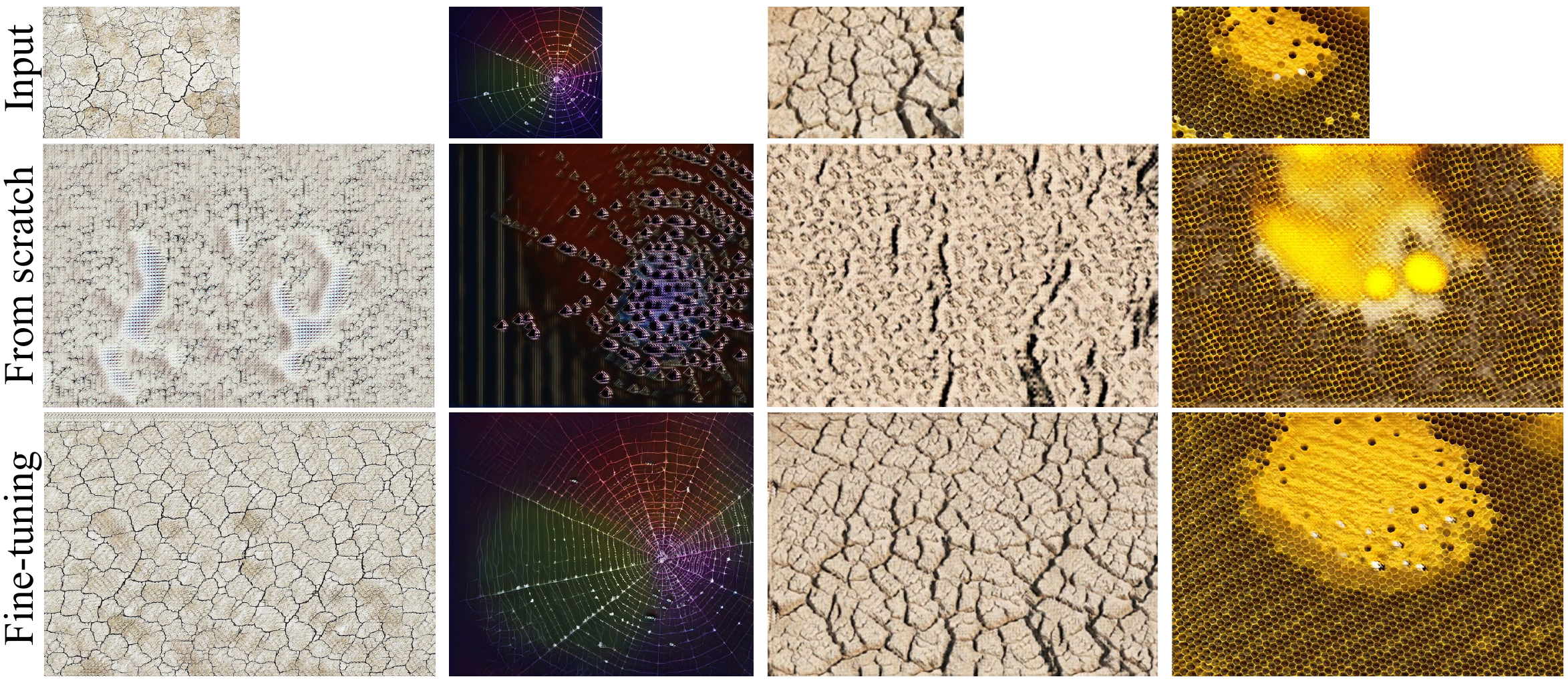}
% 	\vspace{-3ex}
	\caption{Fast expansion on unseen textures.}
    \label{fig:unseen}
    % \vspace{-1ex}
\end{figure}

\vspace{1mm}\noindent\textbf{Extension at the instance level from the same category.} In order to verify the fast expansion strategy at the instance level, we experiment with Internet images for each texture category. We fine-tune a pre-trained model with an unseen image for 900 iterations and compare to a model that is trained from scratch for 900 iterations. As shown in Fig.~\ref{fig:unseen}, training from scratch can hardly capture sufficient texture information and generate reasonable images. 

\begin{figure}
	% \vspace{-0.5cm}
	\centering
	\includegraphics[width=0.49\textwidth]{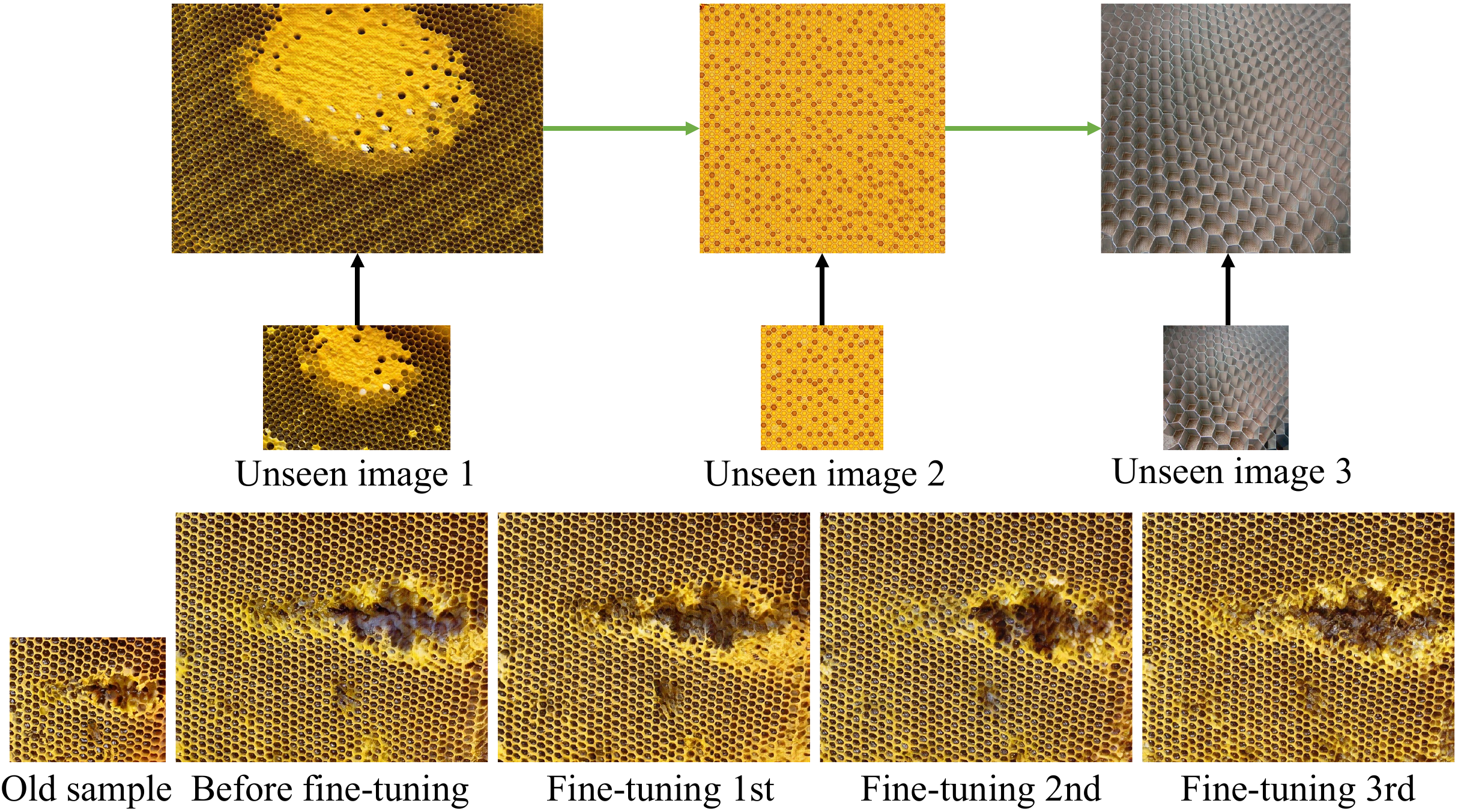}
	\caption{Sequential fine-tuning on the same category.}
    \label{fig:Sequential}
\end{figure}

Besides, we conduct another experiment to verify the stability and generalization of our model. We sequentially fine-tune a well-trained model on three unseen honeycombed textures. That is, when we fine-tune the model on the second texture, the first image is added to the old sample set. Fig.~\ref{fig:Sequential} shows the sequential fine-tuning results of these three textures. It can be seen that although the model has been fine-tuned three times sequentially, it can still cope with old samples well.

In general, STS models usually takes at least two hours to generate a new reasonable result (Tab.~\ref{tab:time}), our strategy is efficient by reducing the time to 5 minutes. Although some MTS methods take less time than ours to expand unseen images, they cannot cope with {\it non-stationary} textures. 
Therefore, this strategy facilitates the stability and generalization of our model, and dramatically reduces the time cost on unseen textures compared to STS models. 

%In general, this extension strategy guarantees the stability and generalization of our model, and dramatically reduces the time on unseen textures compared to STS models.
% \zzhu{Highlight influence.}
% Obviously, the results from fine-tuning are better than those training from scratch. Also note, the fine-tuned model can still cope with the old samples thanks to the replay-based sampling.

\vspace{1mm}\noindent\textbf{Extension at the category level.} Our method can cope with multiple categories jointly within one model through a continual learning-based training strategy. The fourth row of Fig.~\ref{fig:train_strategy} presents the results of different categories from one model under the continual learning strategy. We specify 8,000 epochs for each category, and we successively train five categories considering the time and computational cost. The results obtained by continual learning are noticeably better than random training, though with a slight decrease compared to the category-specific training strategy.

% This experiment proves the effectiveness of the in

% , and both the training strategies are significantly superior to random training. 

% The results obtained by incremental learning also show good effects though with a slight decrease compared to the category-specific training strategy, such as the bricks in the second column and the honeycomb in the third column.

% \begin{figure}
% 	% \vspace{-0.5cm}
% 	\centering
% 	\includegraphics[width=0.48\textwidth]{figs/unseen_test.pdf}
% 	\caption{Fast expansion results. We fine-tune a pre-trained model using a single image for 900 iterations, and train a model from scratch for the same iterations. With a fast fine-tune phase, our method can preserve the global structure and generate better results on unseen images.}
% 	%These lost the structure of the inputs. In the last row, we show the results generated from the fine-tuned model. With a fast fine-tune phase, our method can keep the structure more accurately and generate better results on unseen images.}
%     \label{fig:unseen_test}
% \end{figure}

\subsection{Ablation Study }
\label{chap:ablation}

\begin{figure}
	% \vspace{-0.5cm}
	\centering
	\includegraphics[width=0.48\textwidth]{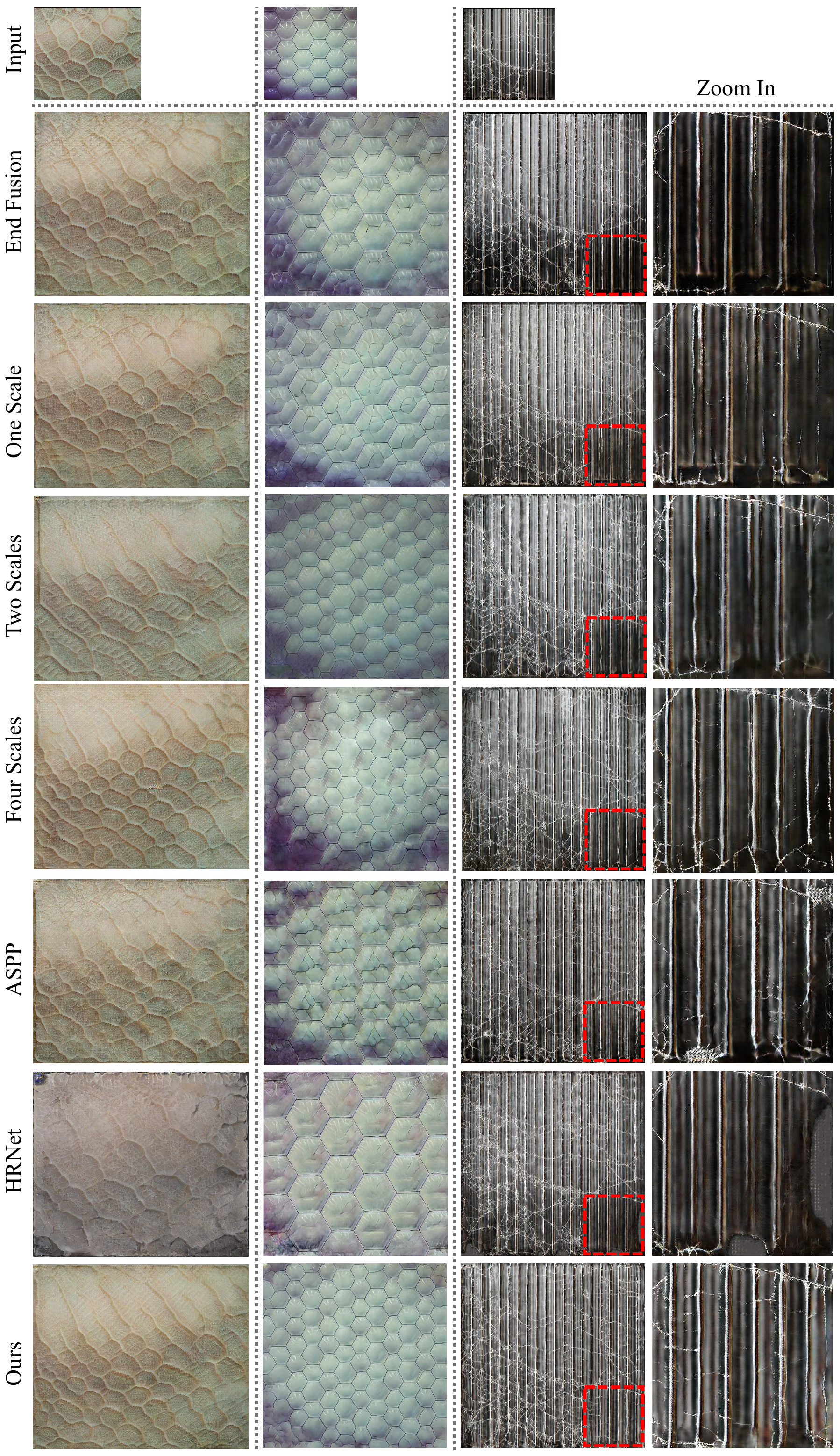}
	\caption{Qualitative results of ablation experiments.}
    \label{fig:ablation}
    % \vspace{-3ex}
\end{figure}
We evaluate several design choices of our generator. Fig.~\ref{fig:ablation} and Tab.~\ref{tab:ablation} demonstrate the qualitative and quantitative comparison results. 
% Obviously, batch normalization will cause serious color distortion. We attribute this reason to information leakage, that is, batch normalization makes information exchange between different samples, including color information. In contrast, using instance normalization is devoid of this problem. 
% More importantly, we can verify the effectiveness of multi-branch design through this set of experiments. 
Note that adding a new branch is essentially introducing a new scale because the down-sampling rate of each branch is different.
From Fig.~\ref{fig:ablation}, the third row (``One Scale'') shows that a single-scale network cannot naturally expand large-scale textures, but simply enlarge them. We observed a similar problem in training STS methods on a category level. Specifically, the ``One-Scale'' setting
of our MTGAN is equivalent to the architecture of Non-stat.~\cite{zhou2018non}. In other words, when training Non-stat. with multiple textures following our category-specific training strategies, Non-Stat. presents the same phenomena as shown in the third row of Fig.~\ref{fig:ablation}.
When adding another branch, the ``Two Scales'' model can conduct preliminary texture expansion with more scale variance. However, the size of hexagons is different, and the color and structure are inconsistent with the original image. 
Furthermore, when adding the third branch (``Ours''), more satisfactory results can be obtained since enough multi-scale information is gathered to deal with such a structure. On the other hand, it is even better for the multi-scale model to handle textures with different structure scales in a single image. For the third column of Fig.~\ref{fig:ablation}, the original image consists of a small-scale cobwebbed texture and a middle-scale railing texture. It is difficult to handle these two textures jointly in a single-scale model, but it can be done in a multi-scale model. 
Besides, adding more scales (``Four Scales'') will not improve the overall performance yet cause the increase of model capacity, according to our experiment of adding four scales (32$\times$ down-sampling). The feature maps of the branch with the highest down-sampling rate are quite small ($4\times 4$), introducing more artifacts during the fusion phase. Therefore, to balance the performance and efficiency, we choose three scales as default. 

\begin{figure}[t]
	% \vspace{-0.5cm}
	\centering
	\includegraphics[width=0.49\textwidth]{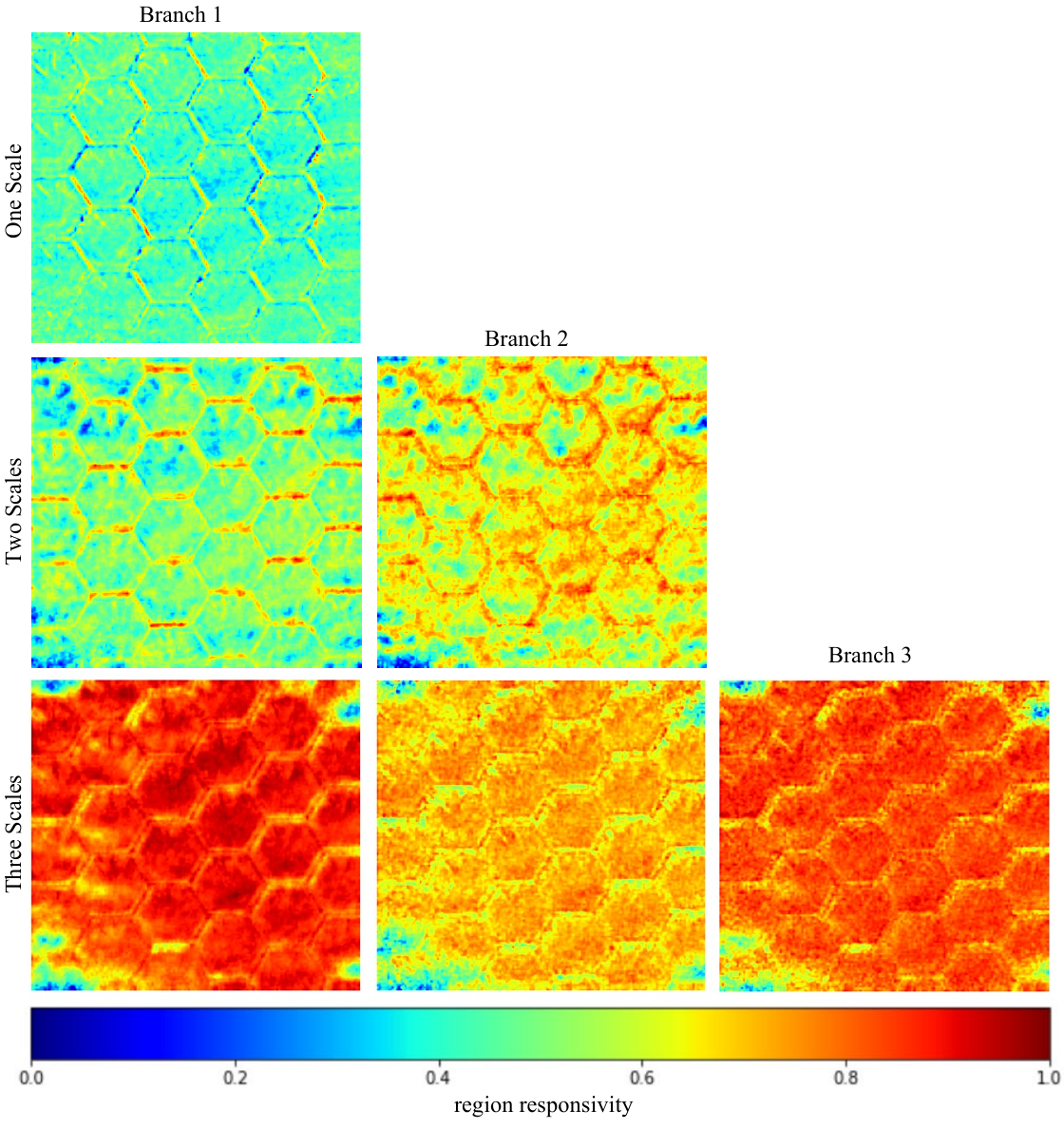}
	\caption{Visualization of feature maps among different branches. }
    \label{fig:visualization}
\end{figure}

% We present the visualization of feature maps among different branches in the supplementary materials to intuitively show the effect of the multi-scale design.
To further intuitively verify the effectiveness of our multi-scale design, we visualize the feature map output by the last ResBlock of each branch. Taking the second column of Fig.~\ref{fig:ablation} as an example, the visualization results are shown in Fig.~\ref{fig:visualization}. When the generator has only one branch (``One scale''), the visualization reflects some focus on local details such as edge information yet shows a rather faint response to global structure. After adding a branch (``Two scales''), ``Branch 2'' pays more attention to the shape information of internal texture elements, such as the shape of a hexagon. ``Branch 2'' integrates this information into the primary branch (``Branch 1'') so that ``Branch 1'' shows more sensitivity to local details and can also capture some shape information. Moreover, after adding the third branch (``Three Scales''), ``Branch 3'' and ``Branch 2'' relatively concentrate on the global structure and spatial arrangement of textures. The primary branch can capture both local details and global structure. With the assistance of our fusion strategy, these three branches work collaboratively to make the model sensitive to both local details and global structure, making non-stationary texture expansion more feasible.

\begin{table}
    \centering
    \small
    \setlength\tabcolsep{5pt}
    \caption{Quantitative results of ablation experiments.}
    \begin{tabular}{|l|c|c|c|c|}
    \hline
        Methods & MSTS $\downarrow$ & SSIM $\uparrow$ & DISTS $\downarrow$  & FID $\downarrow$\\ \hline
        % Batch Norm & 274.55 & 0.0434 & 0.3630 & 146.82    \\ 
        End Fusion & 235.41 & 0.0602 & 0.3150  & 125.69   \\ 
        One Scale & 241.48 & 0.0616 & 0.3144  & 139.48 \\ 
        Two Scales & 230.48 & 0.0619 & 0.3124  & 129.70 \\ 
        Four Scales & 220.72  & 0.0589 & 0.3139  & 126.68 \\ \hline
        ASPP & 234.96  & 0.0597 & 0.3102  & 137.39 \\
        HRNet & 338.60 & 0.0295 & 0.3781  & 181.95 \\ \hline
        Ours (Three Scales) & {\bf216.28} & {\bf 0.0621} & {\bf 0.3046}  & {\bf 123.33} \\ \hline
    \end{tabular}
    \normalsize
    % \vspace{0.1cm}
    \label{tab:ablation}
    % \vspace{-3ex}
\end{table}

% \zzhu{What do you mean by mentioning this experiment?}
% In addition, we set up another branch fusion method, that is, only the last residual blocks will be merged. We find that the results of such an operation are almost the same as that of a single-scale model, because the primary branch does not obtain much valid information from the secondary branches. 

Feature fusion of different branches can be conducted at the end of the residual block chains rather than in every internal residual block. However, we find the result of such fusion manner (``End Fusion'') is almost the same as that of a single-scale model, which implies that the primary branch does not obtain much valid information from secondary branches. This proves the effectiveness of our fusion strategy for our task.

We also compare some representative multi-scale designs by applying ASPP~\cite{chen2017deeplab} module to the one-scale MTGAN and replacing the multi-branch architecture of MTGAN with HRNet~\cite{wang2020deep}. As shown in Fig.~\ref{fig:ablation} and Tab.~\ref{tab:ablation}, both models are inferior to the multi-scale MTGAN for non-stationary multi-texture synthesis. Specifically, HRNet suffers from artifacts and color distortion, and ASPP cannot deal with large-scale textures well. We hypothesize the performance gap is due to what we described in Sec.~\ref{MTGAN}. 

\begin{figure*}[t]
	% \vspace{-0.5cm}
	\centering
	\includegraphics[width=0.90\textwidth]{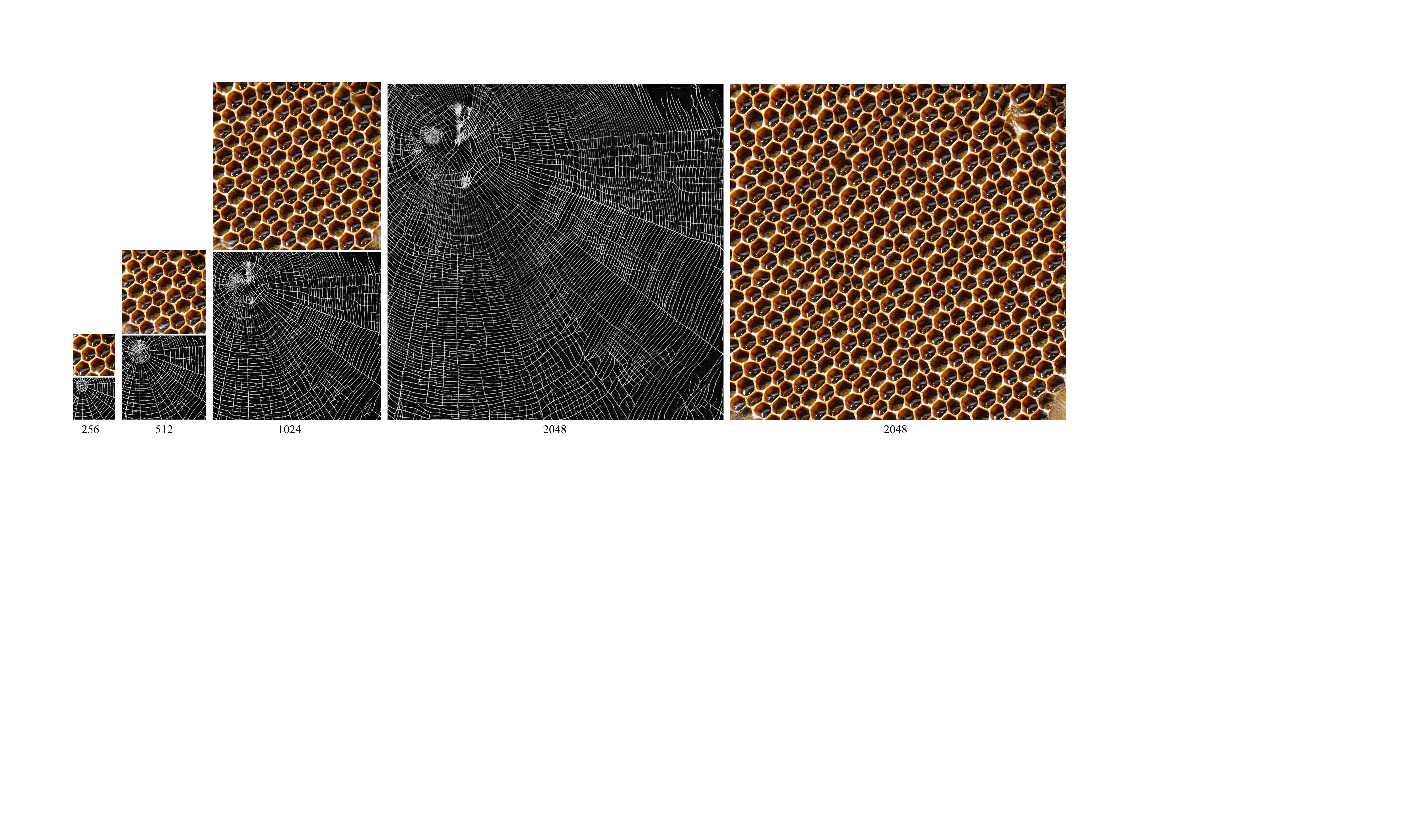}
	\caption{We can synthesize textures of different resolutions by feeding the output of the model as its input circularly. }
    \label{fig:expansion}
\end{figure*}

\begin{figure}
	% \vspace{-0.5cm}
	\centering
	\includegraphics[width=0.45\textwidth]{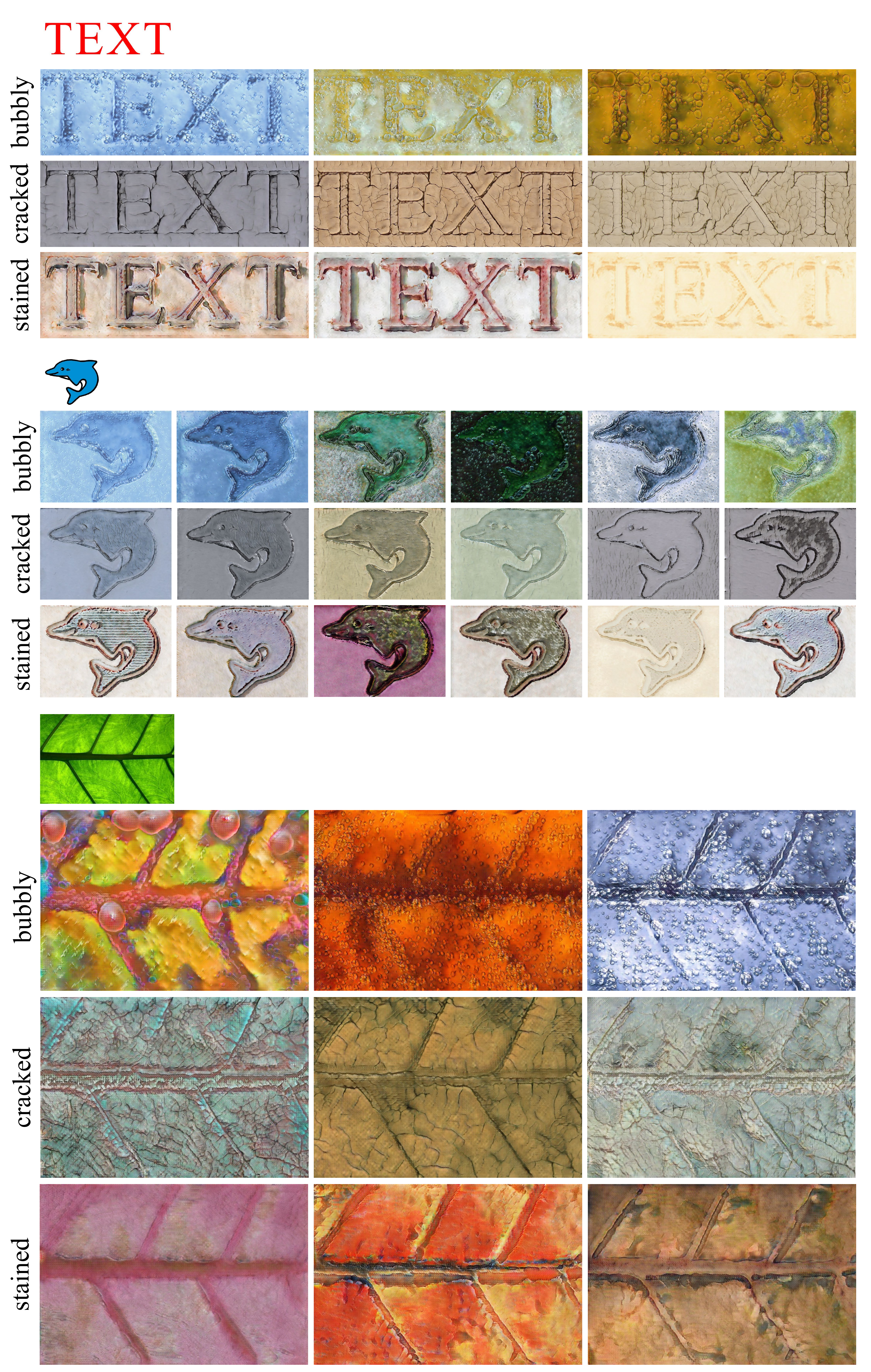}
	\caption{Multi-pattern outputs from three models (bubbly, cracked, stained). }
    \label{fig:multi_modal}
    % \vspace{-3ex}
\end{figure}

\section{Applications}

\noindent\textbf{High-Resolution Texture Synthesis.}~We can obtain textures of different resolutions by repeating the expansion until generating a high-resolution image. That is, we feed the output of the model as its input circularly. In this way, the proposed method synthesizes a $2048\times 2048$ result for each texture, as shown in Fig.~\ref{fig:expansion}. Then, we can crop the large-scale texture to our desired sizes/aspect-ratios to produce wallpapers. Such wallpapers have visual coherence, unlike those rigidly spliced by multiple identical patches with obvious seams in daily life. The details and structure of the texture can be well preserved.

\noindent\textbf{Multi-Pattern Output.}~Training a single model with a class of texture images can potentially empower the model to yield diverse generations. In order to manifest this intriguing property, we feed an image to a trained model and obtain diverse outputs by adding noise or performing some random data augmentations (brightness, contrast, saturation, hue) to the input image. We observe the model can output multiple variations of the same texture category while STS models lack this ability. These results in Fig.~\ref{fig:multi_modal} indicate that our category-specific training strategy brings rich non-stationary texture patterns to the model, and thus can carry out texture-guided multi-pattern image style transfer and WordArt design.

\section{Conclusions}
We focus on non-stationary multi-texture synthesis by designing a systematic scheme including model, training, and evaluation. Extensive experiments demonstrate the effectiveness and superiority of our approach.
Moreover, our model can produce multi-pattern generations of the same category compared to other texture synthesis models. 
% Our future research will explore a more effective continual learning paradigm to extend our model to more categories jointly as well as apply the proposed MTGAN architecture to more image-to-image translation tasks.

The introduced task offers a meaningful research direction to the community. It pushes texture synthesis to practicality, generating more difficult and non-stationary texture images with high fidelity and low time cost. From the application standpoint, these rapidly synthesized non-stationary textures can provide more exquisite and plentiful raw materials for fashion designers, artists, and computer game developers. Academically, the proposed multi-scale architecture may be used for more image-to-image translation tasks, and the category-specific training strategy may offer insight to other class-incremental learning algorithms. We wish the proposed Multi-Scale Texture Similarity could be a general metric to evaluate the texture property in the computer vision community.

\bibliographystyle{IEEEtran}
\bibliography{egbib}

\end{document}